\documentclass[acmtog]{acmart}

\usepackage{cirl}
\usepackage{dipoles}

\setcitestyle{square,nosort}
\setcopyright{rightsretained}
\acmJournal{TOG}
\acmYear{2024} \acmVolume{43} \acmNumber{6} \acmArticle{192} \acmMonth{12}\acmDOI{10.1145/3687914}

\begin{teaserfigure}
	\centering
	\includegraphics{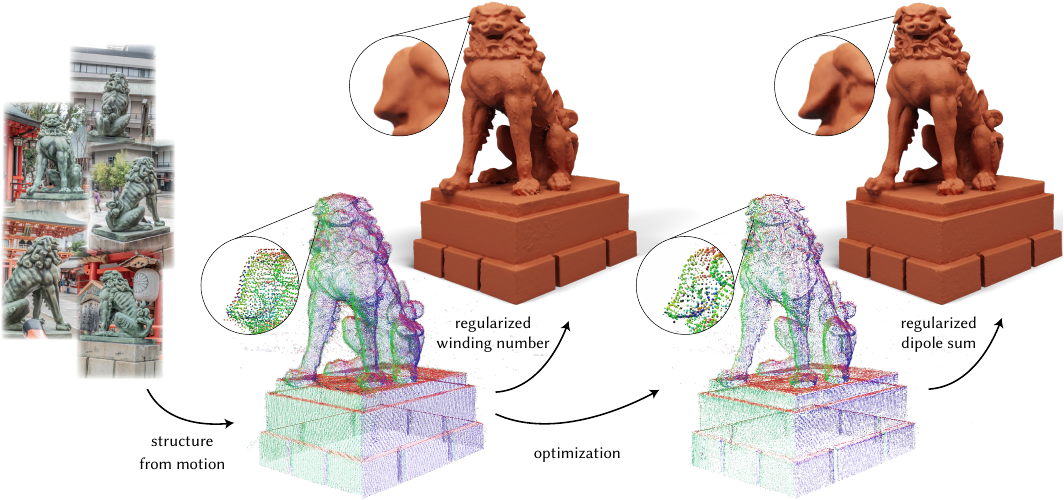}
	 \caption{We introduce the regularized dipole sum, a point-based representation for multi-view 3D reconstruction. This representation can model both implicit geometry and radiance fields using per-point attributes, and supports efficient ray tracing and differentiable rendering, thus facilitating optimization using multi-view images. We initialize our regularized dipole sum representation using the dense point cloud output of a structure from motion procedure (COLMAP). Bootstrapping from this initialization, we use inverse rendering to optimize per-point attributes (visualized in insets as varying point radii), resulting in a higher-quality surface reconstruction. Images are from the  ``Komainu / Kobe / Ikuta-jinja'' dataset by Open Heritage 3D.}
	 \label{fig:teaser}
 \end{teaserfigure}

\begin{document}
\title{3D Reconstruction with Fast Dipole Sums}

\author{Hanyu Chen}
\email{hanyuche@andrew.cmu.edu}
\orcid{0009-0009-0881-0351}

\author{Bailey Miller}
\email{bmmiller@andrew.cmu.edu}
\orcid{0009-0009-0881-0351}

\author{Ioannis Gkioulekas}
\email{igkioule@cs.cmu.edu}
\affiliation{%
  \institution{Carnegie Mellon University}
  \streetaddress{5000 Forbes Ave}
  \city{Pittsburgh}
  \state{PA}
  \postcode{15213}
  \country{USA}
}
\orcid{0000-0001-6932-4642}

\begin{abstract}
We introduce a method for high-quality 3D reconstruction from multi-view images. Our method uses a new point-based representation, the regularized dipole sum, which generalizes the winding number to allow for interpolation of per-point attributes in point clouds with noisy or outlier points. Using regularized dipole sums, we represent implicit geometry and radiance fields as per-point attributes of a dense point cloud, which we initialize from structure from motion. We additionally derive Barnes-Hut fast summation schemes for accelerated forward and adjoint dipole sum queries. These queries facilitate the use of ray tracing to efficiently and differentiably render images with our point-based representations, and thus update their point attributes to optimize scene geometry and appearance. We evaluate our method in inverse rendering applications against state-of-the-art alternatives, based on ray tracing of neural representations or rasterization of Gaussian point-based representations. Our method significantly improves 3D reconstruction quality and robustness at equal runtimes, while also supporting more general rendering methods such as shadow rays for direct illumination.
\end{abstract}

%
%

	\ccsdesc[500]{Computing methodologies~Point-based models}
	\ccsdesc[500]{Computing methodologies~Ray tracing}
%
%
\keywords{Winding number, point-based modeling, inverse rendering}

\maketitle

\section{Introduction}

The emergence of neural rendering methods \citep{tewari2022advances} has led to the widespread adoption of a two-stage pipeline for 3D reconstruction from multi-view images: The first stage uses traditional multi-view geometry methods such as structure from motion \citep{schoenberger2016sfm} to estimate unknown parameters required for the second stage---namely, camera poses. The second stage uses gradient-based optimization and differentiable rendering to optimize a scene representation so that it reproduces the multi-view images---an inverse rendering process. The performance of this pipeline depends critically on the choice of scene representation, motivating the development of various choices (e.g., neural \citep{mildenhall2021nerf,wang2021neus}, grid-based \citep{fridovich2022plenoxels,karnewar2022relu,wu2022voxurf}, hash-encoded \citep{muller2022instant,neus2,li2023neuralangelo}) that offer different tradeoffs between expressive power and computational efficiency.

This paper introduces a new scene representation for multi-view 3D reconstruction, the \emph{regularized dipole sum}. This representation uses tailored kernel-based interpolation of point cloud attributes, to model both the scene geometry (an implicit surface) and scene lightfield (a radiance field). Our representation continues a recent shift towards point-based representations for neural rendering \citep{xu2022point}. In particular, point-based representations using 3D Gaussian kernels have recently gained widespread popularity for both novel-view synthesis tasks \citep{kerbl20233d} and 3D reconstruction \citep{Dai2024GaussianSurfels,huang20242dgs}: The use of Gaussian kernels allows these methods to perform differentiable rendering using image-space rasterization instead of ray tracing, resulting in impressive computational acceleration. At the same time, the use of rasterization precludes combinations of these representations with advanced rendering features such as direct illumination methods (e.g., shadow rays), which rasterization is incompatible with.

By contrast, we design the regularized dipole sum representation to support efficient differentiable rendering with \emph{ray tracing}. Our representation is fundamentally based on the \emph{winding number} for point clouds \citep{barill2018fast}---an approximation to the indicator function of the solid object represented by the point cloud, equal to the sum of Poisson kernels centered at all point cloud locations. The winding number has useful geometric regularization properties \citep{lin2022gauss,lu2018gauss,xu2023globally,metzer2021dipole}, as a jump-harmonic function that approximates the output of robust surface reconstruction algorithms \citep{kazhdan2006poisson}. It is also amenable to efficient computation using fast summation methods \citep{beatson1997short}. Lastly, it can be directly initialized with an optional output of the first-stage structure from motion---a \emph{dense} 3D point cloud of quality approaching that of reconstructions from state-of-the-art neural rendering methods.

The regularized dipole sum generalizes the winding number in several ways that preserve its desirable properties, while also turning it into a point-based representation suitable for inverse rendering applications. As we explain in \cref{sec:regularized}, we use regularized kernels and general per-point attributes, to make this representation compatible with point clouds that are noisy or contain outliers---as point clouds from structure from motion typically do. Then, in \cref{sec:representation}, we show how to use regularized dipole sums to represent not only the geometry, but also the radiance field of a scene. Lastly, in \cref{sec:backprop}, we use fast summation methods to enable efficient computation \emph{and} backpropagation, as needed for inverse rendering.

With the resulting fast dipole sums, we can use ray tracing to optimize a dense point-based representation initialized directly from structure of motion, by simply updating point-based attributes. \Cref{fig:teaser} shows an example use of our approach: Structure from motion \citep{schoenberger2016sfm} produces a dense point cloud that we visualize as a continuous surface using (our regularized generalization of) the winding number. We then use inverse rendering with fast dipole sums to optimize attributes of this point cloud (visualized in the insets), resulting in an improved reconstructed surface. In \cref{sec:experiments}, we evaluate our approach against state-of-the-art neural rendering methods for surface reconstruction, using neural \citep{li2023neuralangelo,neus2} and 3D Gaussian \citep{Dai2024GaussianSurfels} representations. Our experiments show that our approach is both efficient and effective, greatly improving reconstruction quality and robustness at equal runtimes, while additionally supporting rendering with methods such as shadow rays. We provide interactive visualizations and an open-source implementation on the project website.%
\footnote{{\url{https://imaging.cs.cmu.edu/fast_dipole_sums}}}


\section{Related work}\label{sec:prior}

\paragraph{Structure from motion} 3D reconstruction from uncalibrated multi-view images, also known as \emph{structure from motion}, is a classical problem in computer vision \citep{tomasi1990shape,ullman1979interpretation}. It has been the subject of extensive theoretical study \citep{hartley2003multiple} and engineering efforts \citep[Bundler]{snavely2008modeling}---we refer to \citet{ozyecsil2017survey} for a detailed review. Traditional methods attacked this problem primarily by enforcing inter-image geometric consistency, and triangulating correspondences across different images. Mature structure from motion methods \citep{schoenberger2016sfm} can robustly produce sparse point cloud reconstructions from thousands of images \citep{snavely2006photo}. Additionally, such methods provide optional shading-based refinement capabilities \citep{schonberger2016pixelwise} to turn sparse into \emph{dense} point clouds capturing high geometric detail. Lastly, these methods can cover scenes ranging from individual objects \citep{schoenberger2016sfm} to entire cities \citep{agarwal2011building}. We focus on the first setting, and aim to produce high-fidelity \emph{object-level} reconstructions, by directly utilizing and optimizing dense point clouds from structure from motion implementations (\cref{fig:teaser}).

\paragraph{Shading-based refinement and neural rendering} Geometric-only structure-from-motion methods produce point clouds that can have holes in textureless areas where there are no correspondences. They also typically cannot reproduce fine surface details, because they do not exploit shading cues that provide normal information. Shading-aware refinement methods can refine initial structure from motion reconstructions using either simple shading models \citep{dai2017bundlefusion,langguth2016shading,zollhofer2015shading,wu2011high} or complex differentiable rendering procedures \citep{luan2021unified}. However, accounting for shading requires also optimizing for ancillary scene information, such as reflectance and global illumination, resulting in a challenging and ill-posed inverse rendering problem.

Recent neural rendering methods have made tremendous progress towards overcoming these challenges. We refer to  \citet{tewari2022advances} for a detailed review, and discuss only the most relevant works. \Citet{mildenhall2021nerf} tackled multi-view reconstruction problems through the combined use of differentiable volume rendering (implemented through ray tracing), neural field representations for both geometry (implicit surfaces) and global illumination (radiance fields), and structure from motion for pose estimation (COLMAP \citep{schoenberger2016sfm}). Though they initially focused on novel-view synthesis, subsequent methods have adapted this methodological approach for surface reconstruction tasks \citep{yariv2021volume,oechsle2021unisurf,wang2021neus}. Unfortunately, the expressive power neural field representations provide comes with two critical caveats:
\begin{enumerate*}
	\item It introduces a severe computational overhead, resulting in very costly inverse rendering optimization.
	\item It makes it difficult to leverage the 3D reconstruction output of structure from motion in ways more direct and effective than as just regularization during optimization \citep{deng2022depth,fu2022geoneus}.
\end{enumerate*}
We overcome these challenges by developing point-based field representations that are amenable to efficient ray tracing, and can directly optimize dense point clouds from structure from motion.

\paragraph{Geometry and radiance field representations} To alleviate the computational complexity issues due to neural field representations, recent work has made rapid progress towards alternative representations for implicit geometry and radiance fields. Grid-based methods replace neural fields with either dense \citep{karnewar2022relu} or adaptive \citep{fridovich2022plenoxels,wu2022voxurf} grids that are efficient to ray trace \citep{museth2013openvdb} and interpolate, though potentially memory intensive (for dense grids) or difficult to optimize in an end-to-end manner (for adaptive grids). Hash-based methods replace neural fields with multi-resolution hash encodings \citep{muller2022instant,neus2,li2023neuralangelo}, which combine expressive power and efficiency. All these approaches can optionally be combined with shallow (thus more efficient) neural networks that post-process interpolated or encoded features. These approaches overcome computational efficiency issues associated with neural fields, though they still do not provide a way to directly use point clouds available from structure from motion.

Point-based field representations use a point cloud and kernel-based interpolation to compute field quantities needed to express implicit geometry and radiance fields. \Citet{xu2022point} proposed this approach for novel-view synthesis, though their use of complex neural network post-processing of point features still introduces significant computational overhead. \Citet{kerbl20233d} introduced a point-based representation that uses collections of 3D Gaussians to represent both geometry (volumetric density) and radiance. Critically, they also combine this representation with rasterization---through image-space Gaussian splatting---to eliminate the need for costly ray tracing during volume rendering, thus achieving real-time optimization and rendering performance. Though this method originally focused on novel-view synthesis, subsequent works \citep{guedon2023sugar,Dai2024GaussianSurfels,huang20242dgs} have provided extensions for high-fidelity surface reconstruction. Being point-based, these methods can directly leverage 3D information from structure from motion. However, in transitioning from ray tracing to rasterization, they sacrifice generality: for example, rasterization rules out rendering methods such as shadow rays \citep{ling2023shadowneus} (also known as next-event estimation \citet{pharr2023physically}) for rendering direct illumination from know light sources. We contribute a point-based representation that uses \emph{ray tracing}---thus maintaining compatibility with such rendering methods---yet is as efficient as Gaussian splatting methods. Compared to these methods, and to concurrent work \citep{yu2024gaussian} on ray tracing 3D Gaussian representations, our method leverages and extends the advantages afforded by winding number representations \citep{barill2018fast} to produce 3D reconstructions of even higher quality (\cref{sec:experiments}).

\paragraph{Point cloud surface reconstruction} Point-based geometry representations have a long history in computer graphics as methods for reconstructing continuous surfaces (either implicit or, after isosurface extraction \citep{lorensenMarchingCubes}, explicit) from  point clouds. These methods often find use as post-processing of point clouds from structure from motion methods, and thus are robust to imperfections such as noisy points, outlier points, or holes. \Citet{berger2014state} and \citet{huang2024surface} provide detailed reviews. The methods by \citet{fuhrmann2014floating} and \citet{zagorchev2011curvature} use anisotropic Gaussian functions and their derivatives to interpolate scalar fields from point locations, and thus bear a strong similarity to the 3D Gaussian splatting representations we discussed above. \Citet{carr2001reconstruction} use instead more general radial-basis functions for interpolation, combined with fast summation methods \citep{beatson1997short}. Among this extensive family of methods, we build on the point-based winding number representation \citet{jacobson2013robust,barill2018fast,spainhour2024winding}, because of its attractive properties of geometric regularization, robustness, and efficiency---we provide a review in \cref{sec:background}. We generalize winding numbers in \cref{sec:regularized,sec:representation,sec:backprop} into our regularized dipole sum representation, which we use for both implicit geometry and radiance fields. Doing so allows us to achieve efficient inverse rendering of point clouds for high-quality surface reconstruction.

\section{Background}\label{sec:background}

We discuss background on volume rendering with radiance fields for surface reconstruction, and the winding number for point clouds.

\subsection{Inverse volume rendering with radiance fields}\label{sec:rendering}

We follow the methodology introduced by NeRF \citep{mildenhall2021nerf} and represent a 3D scene as a volume comprising two components:
\begin{enumerate*}
	\item an \emph{attenuation coefficient} $\coeff: \R^3 \times \Sph^2 \to \R_{\ge 0}$ representing the scene's geometry; and
	\item a \emph{radiance field} $\radiance: \R^3 \times \Sph^2 \to \R_{\ge 0}^3$ representing the scene's (RGB) lightfield.
\end{enumerate*}
As \citet{miller2023theory} explain, at every scene point $\point\in\R^3$ and direction $\direction \in \Sph^2$, the attenuation coefficient $\coeff\paren{\point,\direction}$ is the probability density that a ray passing through $\point$ along $\direction$ will terminate instantly due to intersection with the scene's geometry. Then, the radiance field $\radiance\paren{\point,\direction}$ is the incident (RGB) global illumination at $\point$ along $\direction$.

This representation allows expressing the RGB intensity (\emph{color}) $\intensity$ captured by a camera ray $\ray_{\camera,\view}\paren{\distancealt}\equiv \camera + \distancealt \view,\, \distancealt\in\R_{\ge 0}$ with origin $\camera$ and direction $\view$ using the (exponential) \emph{volume rendering equation}:
\begin{align}
	\intensity\paren{\camera,\view} = \int_{\distancealt_\mathrm{n}}^{\distancealt_\mathrm{f}} &\exp\paren{-\int_{\distancealt_\mathrm{n}}^\distancealt \coeff\paren{\ray_{\camera,\view}\paren{\distance},\view} \ud \distance} \nonumber \\
	&\qquad\cdot\coeff\paren{\ray_{\camera,\view}\paren{\distancealt},\view} \radiance\paren{\ray_{\camera,\view}\paren{\distancealt}, -\view} \ud \distance,\label{eqn:rendering}
\end{align}
where $\distancealt_\mathrm{n}$ and $\distancealt_\mathrm{f}$ are near and far (resp.) integration limits due to the scene's bounding box. Rasterization approaches \citep{kerbl20233d,zwicker2002ewa,Dai2024GaussianSurfels} approximate \cref{eqn:rendering} by projecting (a point-based representation of) $\coeff$ and $\radiance$ on the image plane, where integration becomes an efficient splatting operation. By contrast, ray tracing approaches approximate $\intensity\paren{\camera,\view}$ with numerical quadrature \citep{max1995optical} using ray samples $\distancealt_\mathrm{n} = \distancealt_0 < \dots < \distancealt_J = \distancealt_\mathrm{f}$:
\begin{equation}\label{eqn:quadrature}
	\intensity\paren{\camera,\view} \approx \sum_{j=1}^{J} \exp\paren{-\sum_{i=1}^{j} \coeff_{i}\Delta_{i}} \paren{1 - \exp\paren{\coeff_{j}\Delta_{j}}} \radiance_{j}
\end{equation}
where at each sample location $\distancealt_j$, $\Delta_j \equiv \distancealt_{j} - \distancealt_{j-1}$, $\coeff_{j} \equiv \coeff\paren{\ray_{\camera,\view}\paren{\distancealt_{j}},\view}$, and $\radiance_{j} \equiv \radiance\paren{\ray_{\camera,\view}\paren{\distancealt_j}, -\view}$. Both approaches are differentiable, allowing propagation of gradients from rendered colors $\intensity$ to the attenuation coefficient $\coeff$ and radiance field $\radiance$. Rasterization is typically faster but also less general than ray tracing, which allows, e.g., using shadow rays to incorporate direct illumination from known light sources \citep{bi2020neural,bi2020deep,hasselgren2022neurips,verbin2024eclipse}.

With this representation at hand, NeRF methods reconstruct a 3D scene from multi-view images by using gradient descent methods \citep{kingma2015adam} to optimize $\coeff$ and $\radiance$, so as to minimize an objective comparing real and rendered images---an \emph{inverse rendering} methodology \citep{loper2014opendr,marschner1998inverse}.

\paragraph{Surface reconstruction} To improve the performance of this methodology in surface reconstruction tasks, prior work \citep{wang2021neus,yariv2021volume,oechsle2021unisurf,miller2023theory} has represented $\sigma$ as an analytic function of a scalar field $\impfunc: \R^3 \to \R$---which we term the \emph{geometry field}---controlling an implicit surface representation of the scene geometry $\implicit_{\impfunc}\subset \R^3$, i.e., $\implicit_{\impfunc} \equiv \curly{\point \in \R^3 : \impfunc\paren{\point} = 0}$ (with the convention that points where $\impfunc\paren{\point} < 0$ are interior points).

We adopt the representation by \citet{miller2023theory},%
\footnote{This representation is similar to NeuS \citep{wang2021neus}, except enforces reciprocity and uses a sigmoid corresponding to a Gaussian process. In our experiments we found that these changes result in significantly improved performance.}
%
which first defines a \emph{vacancy function} in terms of $\impfunc$:
\begin{equation} \label{eqn:vacancy}
	\vacancy\paren{\point} \equiv \cdfunc\paren{\scale\impfunc\paren{\point}},
\end{equation}
where $\scale > 0$ is a user-defined scale factor, and $\cdfunc: \R \to \bracket{0,1}$ is a \emph{sigmoid function} \citep{han1995influence,glorot2011deep} equal to the cumulative distribution function of a standard normal distribution. Thus $\vacancy$ equals $\nicefrac{1}{2}$ when $\impfunc = 0$ (points on the surface $\implicit_{\impfunc}$), approaches 1 as $\impfunc$ increases (exterior points), and 0 as $\impfunc$ decreases (interior points). \citet[Equation (12)]{miller2023theory} relate $\coeff$ to $\vacancy$ as :%
\begin{equation}\label{eqn:surface_density}
	\coeff\paren{\point, \direction} \equiv \frac{\abs{\direction \cdot \nabla\vacancy\paren{\point}}}{\vacancy\paren{\point}}.
\end{equation}
Then, reconstruction uses the above inverse rendering methodology to optimize (through the differentiable \cref{eqn:surface_density,eqn:vacancy}) the geometry field $\impfunc$  instead of the attenuation coefficient $\coeff$.

\paragraph{Structure-from-motion initialization} Inverse rendering requires knowledge of camera locations $\camera$ and poses $\view$ for each image in a multi-view dataset, to render images with \cref{eqn:rendering}. Inverse rendering methods typically include an initialization stage that uses structure from motion \citep{ozyecsil2017survey} to estimate this camera information. The initialization stage additionally outputs a \emph{sparse} point cloud reconstruction of the 3D scene, and \citet{deng2022depth} show that the final reconstruction can improve by leveraging this output during subsequent inverse rendering optimization.

Modern structure-from-motion methods such as COLMAP \citep{schoenberger2016sfm} provide an optional refinement process \citep{schonberger2016pixelwise} that outputs a \emph{dense} point cloud reconstruction. Inverse rendering methods skip this refinement process during initialization, despite the fact that:
\begin{enumerate*}
	\item it produces reconstructions of quality competitive with or often even better than what inverse rendering achieves \citep{wang2021neus} (we revisit this point in \cref{sec:experiments}); and
	\item its runtime is a lot shorter than the runtime of inverse rendering.
\end{enumerate*}
Thus, using this refinement process during initialization and leveraging leveraging its dense point cloud output for inverse rendering could help greatly improve performance in terms of both reconstruction quality and computational efficiency.

\begin{figure}
	\centering
	\includegraphics[width=\linewidth]{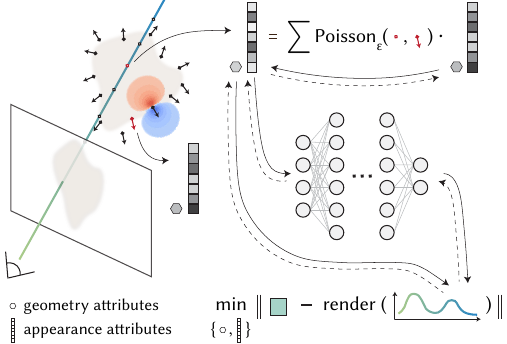}
	\put(-95pt,125pt){\putbox{\small \textsf{Eqs.~\labelcref{eqn:dipole_sum},~\labelcref{eqn:appearance_features}}}}
	\put(-38pt,55pt){\putbox{\small \textsf{Eq.~\labelcref{eqn:radiance_field}}}}
	\put(-110pt,38pt){\putbox{\small \textsf{Eqs.~\labelcref{eqn:geometry_field},~\labelcref{eqn:vacancy},~\labelcref{eqn:surface_density}}}}
	\put(-89pt,5pt){\putbox{\small \textsf{Eq.~\labelcref{eqn:quadrature}}}}
	\caption{Overview of our method. During forward rendering (indicated by solid arrows), at each sample location along a ray, we interpolate geometry and appearance attributes from a point cloud through a fast primal dipole sum query. We pass appearance attributes through a shallow MLP to predict colors, and use geometry attributes to compute attenuation coefficients. We integrate along the ray to compute the rendered color and minimize the \(L^1\)-loss between the rendered and ground truth colors. During backpropagation (indicated by dashed arrows), we optimize geometry and appearance attributes of the point cloud through a fast adjoint dipole sum query.}
	\label{fig:overview}
	\vspace{-1em}
\end{figure}

\paragraph{Our contribution} Within this context, we develop a point-based representation for the geometry field $\impfunc$ and radiance field $\radiance$ that:
\begin{enumerate}[leftmargin=*]
	\item facilitates \emph{fast} ray tracing, enabling efficient inverse rendering (like rasterization) without sacrificing generality (shadow rays);
	\item leverages \emph{dense} point cloud outputs from structure-for-motion initialization, optimizing only per-point attributes and a shallow multi-layer perceptron (MLP) during inverse rendering;
	\item reconstructs high quality surfaces by implicitly enforcing geometric regularization (e.g., harmonicity).
\end{enumerate}
\Cref{fig:overview} overviews our overall method.

\subsection{Winding number}\label{sec:winding}

We derive our point-based representation as a generalization of the \emph{winding number}, which we discuss next.

\paragraph{Continuous surfaces} We first consider the winding number for a continuous surface $\boundary \subset \R^3$.
%
%
Among its many equivalent definitions \citep{feng2023winding}, we use that as a Laplacian \emph{double layer potential with unit moment}, which facilitates the generalizations we consider in \cref{sec:regularized}. Then, the \emph{winding number} $\wn : \R^3 \to \R$ equals:
\begin{equation}\label{eqn:wn}
	\wn\paren{\point} \equiv \int_{\boundary} \poisson\paren{\point, \pointalt} 1 \surfMeasure\paren{\pointalt},\quad \poisson\paren{\point,\pointalt} \equiv \frac{1}{4\pi}\frac{\normal\paren{\pointalt}\cdot\paren{\pointalt-\point}}{\norm{\pointalt - \point}^3}.
\end{equation}
Here, $\normal\paren{\pointalt}$ is the outward normal vector at point $\pointalt \in \boundary$, and $\poisson: \R^3\times\R^3 \to \R$ is the \emph{free-space Poisson kernel} for the Laplace equation. We make explicit the factor $1$ in \cref{eqn:wn}, for reasons we will explain in \cref{sec:regularized}. The scalar field $\wn$ is jump-harmonic and, when the surface $\boundary$ is watertight, equals its binary \emph{indicator function} (a fact known as Gauss' lemma \citep[Proposition 3.19]{folland2020introduction}):
\begin{equation}\label{eqn:gauss}
	\wn\paren{\point} = \begin{cases}
		1, & \point \text{ inside } \boundary, \\[-1pt]
		0, & \point \text{ outside } \boundary, \\[-1pt]
		\nicefrac{1}{2}, & \point \text{ on } \boundary.
	\end{cases}
\end{equation}

\paragraph{Point clouds} We next consider the winding number for an \emph{oriented} point cloud $\pcloud \equiv \curly{\paren{\pc_m,\pcnormal_m,\pcarea_m}}_{m=1}^M$, where for each $m$ we assume that:
\begin{enumerate*}
	\item the point $\pc_m$ is a sample from an underlying surface $\boundary$;
	\item the vector $\pcnormal_m$ is the outward normal of $\boundary$ at $\pc_m$; and
	\item the scalar $\pcarea_m$ is the geodesic Voronoi area on $\boundary$ of $\pc_m$, i.e., the area of the subset of $\boundary$ where points are closer (in the geodesic distance sense) to $\pc_m$ than any other point in $\pcloud$.
\end{enumerate*}
We use the dense point cloud from structure-from-motion initialization, which provides points $\pc_m$ and normals $\pcnormal_m$, and we estimate area weights $\pcarea_m$ as in \citet{barill2018fast}.
Then, \citet{barill2018fast} generalize the winding number to point clouds using a discretization of the double layer potential \labelcref{eqn:wn}.%
\footnote{Throughout we use bars and tildes to indicate correspondences between quantities involving continuous surface integrals and their point-cloud approximations (resp.).}
\begin{myTitledBox}{Winding number for an oriented point cloud}
	For an oriented point $\pcloud \equiv \curly{\paren{\pc_m,\pcnormal_m,\pcarea_m}}_{m=1}^M$, its \emph{winding number} $\pcwn : \R^3 \to \R$ is the scalar field:
	\begin{equation}\label{eqn:pcwn}
		\pcwn\paren{\point} \equiv \sum_{m=1}^M \pcarea_m \poisson\paren{\point, \pc_m} 1 = \sum_{m=1}^M \frac{\pcarea_m}{4\pi}\frac{\pcnormal_m\cdot\paren{\pc_m-\point}}{\norm{\pc_m-\point}^3} 1.
	\end{equation}
\end{myTitledBox}
%

\paragraph{Winding number as a geometry field} Though $\pcwn$ is not a binary scalar field (unlike its continuous counterpart $\wn$), its behavior is still suggestive of the continuous surface $\boundary$ underlying $\pcloud$: As \citet{barill2018fast} show, it approaches $\nicefrac{1}{2}$ at points near the continuous surface $\boundary$ underlying $\pcloud$, increases towards its interior, and decreases towards its exterior. Thus at first glance, it appears we can use it to represent a geometry field for inverse rendering (\cref{sec:rendering}) as:
\begin{equation}\label{eqn:wn_field}
	\pcimpfunc\paren{\point} \equiv \frac{1}{2}-\pcwn\paren{\point},
\end{equation}
corresponding to an implicit surface $\pcshape \equiv \curly{\point \in \R^3: \pcimpfunc\paren{\point} = 0}$. This representation provides several critical advantages:
\begin{itemize}[leftmargin=*]
	\item[\cmark] {\it Accurate approximation.} $\pcshape$ provides an approximation to $\boundary$ that becomes exact as point density becomes infinite, and degrades gracefully as the number of points $M$ decreases.
	\item[\cmark] {\it Geometric regularization.} $\pcimpfunc$ is imbued with regularity properties that provide \emph{geometric regularization}. It is \emph{jump-harmonic}, and thus of a smooth nature that has proven useful for geometric optimization tasks \citep{peng2021shape,lipman2021phase}. It is also related to robust geometric representations \citep{kazhdan2006poisson,belyaev2013signed} and interpolation schemes \citep{floater2005mean,ju2005mean} that have found great success in reconstruction applications. We elaborate on these relationships in \cref{sec:regularization,sec:representation}.
	\item[\cmark] {\it Direct initialization.} $\pcimpfunc$ can be directly computed using the point cloud from structure-from-motion initialization. Point queries for $\pcwn$, and thus $\pcimpfunc$, use only the point cloud attributes, and do not require meshing or a proxy data structure (e.g., grid or neural).
	\item[\cmark] {\it Fast queries.} Such point queries, \emph{and backpropagating through them}, can be made efficient with logarithmic complexity $\bigO\paren{\log M}$ relative to point cloud size $M$, as we explain in \cref{sec:backprop}. Thus, $\pcimpfunc$ lends itself to efficient ray tracing (which requires multiple point queries along each viewing ray (\cref{eqn:rendering})), even when working with dense point clouds from structure from motion.
\end{itemize}
\Citet{barill2018fast} further discuss the benefits of the winding number $\pcwn$ versus other point-based surface representations. At the same time, $\pcwn$, and thus $\pcimpfunc$, have critical shortcomings that make them unsuitable for direct use for inverse rendering:
\begin{itemize}[leftmargin=*]
	\item[\xmark] {\it Numerical instability.} The Poisson kernel $\poisson\paren{\point,\pointalt}$ is singular as $\point \to \pointalt$. The singularity makes the surface $\pcshape$ numerical algorithms interface with---e.g., during ray tracing \citep[Section 4.3]{Gillespie:2024:RTH} or isosurface extraction \citep[Section 3 \& Figure 9]{barill2018fast}---inaccurate and numerically unstable near $\pcloud$. These numerical issues hinder inverse rendering performance (e.g., due to rays passing near or through points in $\pcloud$, \cref{sec:experiments}).
	\item[\xmark] {\it Exact interpolation.} The singularity makes the implicit surface $\pcshape$ an \emph{exact interpolant} of the point cloud $\pcloud$. Exact interpolation is undesirable when working with imperfect point clouds with \emph{noisy} point locations \citep[Section 9]{barill2018fast}, such as those from structure-from-motion initialization.
	\item[\xmark] {\it Outlier sensitivity.} Such point clouds typically also suffer from outlier points (e.g., due to incorrect correspondences) and inaccurate or incorrectly oriented normals. As $\pcwn$ weighs all points equally, it is can be very sensitive to such defects.
\end{itemize}
We explain how to overcome these shortcomings in the next section.

\section{Regularized dipole sums}\label{sec:regularized}

We introduce a generalization of $\pcwn$ in \cref{eqn:pcwn} that facilitates point-based representations in inverse rendering for both the geometry field $\impfunc$ and, as we explain in \cref{sec:radiance}, the radiance field $\radiance$. Our generalization changes \cref{eqn:pcwn} by replacing:
\begin{enumerate*}
	\item singular with non-singular kernels (\cref{sec:regularization}); and
	\item unit with variable per-point weights (\cref{sec:moment}).
\end{enumerate*}
We also present two technical results (\cref{pro:psr,pro:expected_winding_number}) that lend theoretical support to our generalization, by relating it to Poisson surface reconstruction \citep{kazhdan2006poisson} and stochastic point clouds (resp.).

\subsection{Regularization}\label{sec:regularization}

To overcome shortcomings due to the singular Poisson kernel $\poisson$, we turn to \emph{regularization} schemes common in methods for the simulation of linear partial differential equations (e.g., method of fundamental solutions, boundary element method \citep[Section 2.2]{chen2024fast}). These methods use regularization to address numerical issues arising from singular potential kernels analogous to the issues we encounter in inverse rendering.

\begin{figure}
	\centering
	\includegraphics[width=\linewidth]{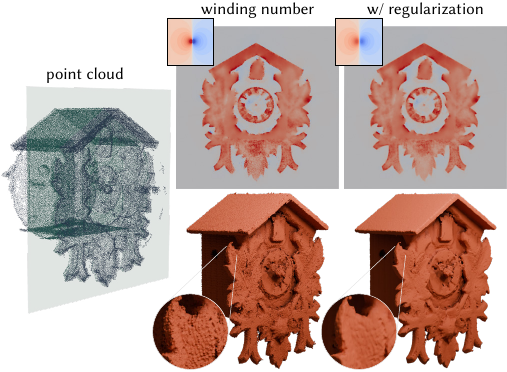}
	\caption{Using the original and regularized winding number fields on the unoptimized point cloud (left) for the BlendedMVS \textsc{clock} scene. The top row shows planar slices of the two fields: The original winding number is very noisy near point cloud locations due to the singular Poisson kernel, whereas the regularized winding number is much smoother. The insets visualize the singular and regularized kernels. The bottom row shows meshes extracted from the two fields using marching cubes: The original winding number results in strong artifacts, which the regularized winding number fixes.}
	\label{fig:regularized}
	\vspace{-1em}
\end{figure}

A common regularization scheme \citep{beale2016simple,cortez2001method,cortez2005method}%
\footnote{An alternative to regularization is to ``desingularize'' the Poisson kernel by introducing a small cutoff in the denominator \citep{lu2018gauss,lin2022gauss}. We found that this approach results in worse performance in inverse rendering experiments (\cref{sec:experiments}), corroborating the arguments of \citet{cortez2001method} in favor of regularization.}
starts from the definition of the Poisson kernel through the Laplacian \emph{Green's function} $\green: \R^3 \times \R^3 \to \R$:
\begin{align}
	\poisson\paren{\point,\pointalt} &\equiv \normal\paren{\pointalt} \cdot \nabla_{\pointalt} \green\paren{\point, \pointalt}, \label{eqn:poisson}\\
	\text{ where $\green$ satisfies: } \Delta_\point \green\paren{\point, \pointalt} &= \delta\paren{\point - \pointalt},\label{eqn:green}
\end{align}
and $\delta$ is the Dirac delta distribution in $\R^3$. Regularization proceeds by replacing $\delta$ with a \emph{nascent delta function}, that is, a function $\nascent_\varepsilon\paren{\point - \pointalt}$ satisfying $\lim_{\varepsilon \to 0} \nascent_\varepsilon\paren{\point - \pointalt} = \delta\paren{\point - \pointalt}$. Then, we can define the \emph{regularized Green's function} $\green_\varepsilon$ and \emph{regularized Poisson kernel} $\poisson_\varepsilon$ exactly analogously to \cref{eqn:poisson,eqn:green}:
\begin{align}
	\poisson_\varepsilon\paren{\point,\pointalt} &\equiv \normal\paren{\pointalt} \cdot \nabla_{\pointalt} \green_\varepsilon\paren{\point, \pointalt}, \label{eqn:poisson_regularized}\\
	\text{ where $\green_\varepsilon$ satisfies: } \Delta_\point \green_\varepsilon\paren{\point, \pointalt} &= \nascent_\varepsilon\paren{\point - \pointalt}.\label{eqn:green_regularized}
\end{align}
It follows that $\lim_{\varepsilon \to 0} \green_\varepsilon = \green$ and $\lim_{\varepsilon \to 0} \poisson_\varepsilon = \poisson$. A common choice of nascent delta function is the Gaussian function:
\begin{equation}\label{eqn:nascent}
	\phi_\varepsilon\paren{\point - \pointalt} \equiv \frac{1}{\varepsilon\sqrt{2\pi}}\exp\paren{-\frac{\norm{\point - \pointalt}^2}{2 \varepsilon^2}}.
\end{equation}
The corresponding regularized Poisson kernel is \citep{beale2016simple}:
\begin{equation}
	\poisson_\varepsilon\paren{\point,\pointalt} \equiv \regul\paren{\frac{\norm{\pointalt-\point}}{\varepsilon}}\poisson\paren{\point,\pointalt},\label{eqn:regularized_poisson}
\end{equation}
where $\regul\paren{t} \equiv \erf\paren{t} - \nicefrac{2 t}{\sqrt{\pi}} \exp\paren{-t^2}$. Unlike $\poisson$, $\poisson_\varepsilon$ is not singular, as $\poisson_\varepsilon\paren{\pointalt,\pointalt} = 3^{-1} \varepsilon^{-3} \pi^{-\nicefrac{3}{2}}$ is finite for $\varepsilon > 0$. The parameter $\varepsilon$ controls the trade-off between regularization (restricting how fast $\poisson_\varepsilon\paren{\point,\pointalt}$ increases as $\norm{\pointalt-\point}\to 0$) and bias (bounding the difference $\poisson_\varepsilon-\poisson$). We can use $\poisson_\varepsilon$ to generalize \cref{eqn:pcwn} as follows.
\begin{myTitledBox}{Regularized winding number for an oriented point cloud}
	For an oriented point $\pcloud \equiv \curly{\paren{\pc_m,\pcnormal_m,\pcarea_m}}_{m=1}^M$, and a regularization parameter $\varepsilon \ge 0$, its \emph{regularized winding number} $\regpcwn : \R^3 \to \R$ is the scalar field:
	\begin{equation}\label{eqn:regularized_pcwn}
		\regpcwn\paren{\point} \equiv \sum_{m=1}^M \pcarea_m \poisson_\varepsilon\paren{\point, \pc_m} 1, 
	\end{equation}
	where $\poisson_\varepsilon$ is the regularized Poisson kernel in \cref{eqn:regularized_poisson}.
\end{myTitledBox}
Unlike $\pcwn$, $\regpcwn$ can be robustly evaluated arbitrarily close to points in $\pcloud$, and its $\nicefrac{1}{2}$-level set does \emph{not} exactly interpolate those points. Thus, it escapes the first two shortcomings we identified at the end of \cref{sec:winding}. During inverse rendering (\cref{sec:inverse}), we can use additional image-based losses to penalize large deviations between the level set and $\pcloud$, while also allowing for inexact interpolation to account for noise. \Cref{fig:regularized} uses the dense point cloud output from dense structure-from-motion initialization for a scene from the BlendedMVS dataset \citep{BlendedMVS} to compare:
\begin{enumerate*}
	\item 2D slices of the original and regularized winding number fields, and
	\item meshed isosurfaces extracted from them using marching cubes \citep{lorensenMarchingCubes}.
\end{enumerate*}
Simply using the regularized winding number on this initial point cloud, without any training, already produces high-quality meshes comparable to those from state-of-the-art inverse rendering methods, as we quantify in \cref{sec:experiments}.

\paragraph{Relationship to Poisson surface reconstruction} We remark on a relationship between the regularized winding number $\regpcwn$ in \cref{eqn:regularized_pcwn}, and \emph{Poisson surface reconstruction (PSR)} \citep{kazhdan2006poisson,kazhdan2013screened}; this relationship highlights the useful geometric regularization properties of $\regpcwn$, and its generalization in \cref{eqn:dipole_sum}.
Like \cref{eqn:regularized_pcwn}, PSR uses an oriented point cloud to compute a scalar field that approximates the continuous winding number \labelcref{eqn:wn} for the underlying surface $\boundary$. This scalar field has proven to enable robust surface reconstruction, thanks to the regularity properties of the Poisson equation used to compute it. As a result, PSR has become a workhorse for point-based surface reconstruction \citep{berger2014state,huang2024surface}. Unfortunately, using PSR in inverse rendering is prohibitively expensive, as querying (and differentiating) its scalar field output requires constructing a grid and performing a global Poisson solve operation \citep{peng2021shape}. However, we prove in \cref{app:psr} the following result.

\begin{prp}[label={pro:psr}]{Poisson surface reconstruction}{psr}
	The regularized winding number $\regpcwn$ is the solution to the Poisson equation of \citet{kazhdan2006poisson}.
\end{prp}

\Cref{pro:psr} shows that we can use the regularized winding number $\regpcwn$ in \cref{eqn:regularized_pcwn} (and its generalization in \cref{eqn:dipole_sum}) as a geometry representation for inverse rendering that has the same regularity and robustness properties as PSR, while remaining efficient to render and differentiate (\cref{sec:backprop}). \Citet[Section 1.1.2]{feng2023winding} and \citet[Section 2.1]{barill2018fast} have previously discussed the relationship between the \emph{non-regularized} winding number $\pcwn$ in \cref{eqn:wn} and PSR. However, the two are equivalent only \emph{asymptotically}, at the limit of zero-variance Gaussian blurring of normals in PSR \citep[Equation (2)]{kazhdan2006poisson}. By contrast, the equivalence \cref{pro:psr} is \emph{exact} for all blur variances---intuitively, the Gaussian blurring of normals in PSR is equivalent to the Gaussian regularization of the Green's function and Poisson kernel in \cref{eqn:poisson_regularized,eqn:green_regularized}. The need to incorporate Gaussian regularization helps resolve, both theoretically and in practice (\cref{fig:regularized}) performance discrepancies between the winding number and PSR in, e.g., isosurface extraction \citep[Section 3 \& Figure 9]{barill2018fast}.

\subsection{Variable moment}\label{sec:moment}

To overcome shortcomings due to outlier points and inaccurate normals, we modify \cref{eqn:regularized_pcwn} to use variable per-point weights---thus allowing to deemphasize outliers. This modification yields a point-based representation suitable for both the geometry field and radiance field (\cref{sec:radiance}) in inverse rendering.

To this end, we can replace the unit moment in \cref{eqn:wn} with a \emph{variable moment} $\impfeat : \boundary \to \R$. Its corresponding \emph{double layer potential} $\dl{\impfeat} : \R^3 \to \R$ is the scalar field \citep[Section 3.C]{folland2020introduction}:
\begin{equation} \label{eqn:double_layer_potential}
	\dl{\impfeat}\paren{\point} \equiv \int_{\boundary} \poisson\paren{\point, \pointalt} \impfeat\paren{\pointalt} \surfMeasure\paren{\pointalt}.
\end{equation}
%
For any sufficiently smooth moment $\impfeat$, $\dl{\impfeat}$ is \emph{jump-harmonic} \citep{krutitskii2001jump}: it satisfies Laplace's equation at $\point\in \R^3\setminus\boundary$, and has a jump discontinuity equal to $\impfeat$ at $\point\in\boundary$, analogously to \cref{eqn:gauss} for $\wn$.

\begin{figure*}
	\centering
	\includegraphics[width=\textwidth]{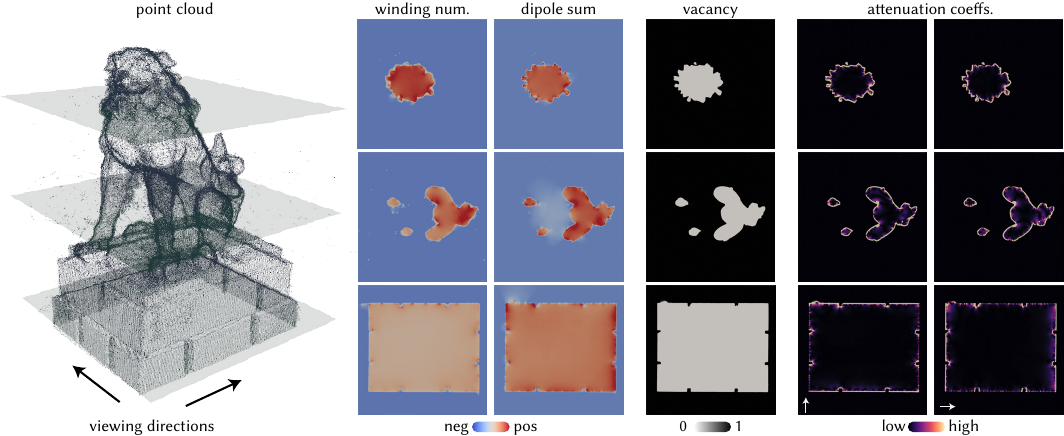}
	\caption{We visualize on the teaser scene geometry-related field quantities that we use for inverse rendering. From left to right: the initial geometry field with unit geometry attributes (equal to the regularized winding number in \cref{eqn:regularized_pcwn}), the optimized geometry field with learned geometry attributes (\cref{eqn:geometry_field}), the optimized vacancy field (\cref{eqn:vacancy}), and attenuation coefficients computed (\cref{eqn:surface_density}) along two different viewing directions.}
	\label{fig:fields}
	\vspace{-1em}
\end{figure*}

Using the moment values on the point cloud, $\impfeat_m \equiv \impfeat\paren{\pc_m}, \pc_m \in \pcloud$, and the regularized kernel $\poisson_\varepsilon$ to circumvent singularity issues, we arrive at a regularized point-cloud approximation of \cref{eqn:double_layer_potential}.%
\footnote{Following \citet[Section 3.1]{barill2018fast} and \citet[Section 2]{gotsman2024linear}, we use the term \emph{dipole} because the Poisson kernel $\poisson\paren{\point,\pointalt}$ equals the electric potential of a dipole centered at $\pointalt$ and polarized in the direction of $\normal\paren{\pointalt}$.}
\begin{myTitledBox}{Regularized dipole sum for an oriented point cloud}
	For an oriented point $\pcloud \equiv \curly{\paren{\pc_m,\pcnormal_m,\pcarea_m}}_{m=1}^M$, a regularization parameter $\varepsilon \ge 0$, and a moment function with point samples $\impfeat_m, m=1,\dots,M$, the corresponding \emph{regularized dipole sum} $\regpcwn : \R^3 \to \R$ is the scalar field:
	\begin{equation}\label{eqn:dipole_sum}
		\regpcdl{\impfeat}\paren{\point} \equiv \sum_{m=1}^M \pcarea_m \poisson_\varepsilon\paren{\point, \pc_m} \impfeat_m.
	\end{equation}
	where $\poisson_\varepsilon$ is the regularized Poisson kernel in \cref{eqn:regularized_poisson}.
\end{myTitledBox}
The point-cloud winding number and its regularized form in \cref{eqn:pcwn,eqn:regularized_pcwn} are special cases, i.e., $\pcwn = \pcdl{\onedirichlet}_{0}$ and $\regpcwn = \pcdl{\onedirichlet}_{\varepsilon}$. The regularized dipole sum maintains the advantages of the point-cloud winding number we listed in \cref{sec:winding}, while addressing its shortcomings. We can thus treat $\impfeat_m, \, m=1,\dots,M$ as a \emph{learnable} per-point \emph{geometry attribute}, and use its corresponding dipole sum to define a point-based representation for the geometry field:
\begin{equation}\label{eqn:geometry_field}
	\impfunc\paren{\point} \equiv \frac{1}{2} - \regpcdl{\impfeat}\paren{\point},
\end{equation}
We then convert $\impfunc$ to a vacancy $\vacancy$ and attenuation coefficient $\coeff$ using \cref{eqn:vacancy,eqn:surface_density}. We initialize $\regpcdl{\impfeat}$ to equal the regularized winding number $\regpcwn$ in \cref{eqn:regularized_pcwn} by using initial geometry attribute values  $\impfeat_m=1$, which we then optimize during inverse rendering to update the scene geometry (\cref{sec:inverse}). \Cref{fig:fields} visualizes these initial and optimized fields on the teaser scene. Compared to the winding number, allowing non-unit values for $\impfeat_m$ during the inverse rendering process serves two goals:
\begin{enumerate*}
	\item It allows the process to diminish the influence of point cloud outliers, by decreasing their geometry attribute $\impfeat_m$. The point cloud insets in \cref{fig:teaser} visualize this effect, by scaling point radii by their optimized geometry attribute.
	\item It allows the process to modify the scene geometry (e.g., to correct noisy point locations or holes in textureless regions) \emph{without} changing the point locations $\pc_m$ in $\pcloud$; as we explain in \cref{sec:backprop_details}, fixing point locations facilitates faster inverse rendering.
\end{enumerate*}

\paragraph{Stochastic point cloud interpretation} Our generalization of the winding number $\pcwn$ in \cref{eqn:wn} into the regularized dipole sum $\regpcdl{\impfeat}$ in \cref{eqn:dipole_sum} was motivated by the need for improved robustness when working with imperfect point clouds that include noisy and outlier points. We can model such a point cloud as \emph{stochastic}, treating both point locations and normals as random variables. The winding number of such a stochastic point cloud is itself a random variable. We then prove the following relationship between this random variable and the regularized dipole sum.

\begin{prp}[label={pro:expected_winding_number}]{Stochastic point cloud}{spc}
	We assume that $\pcloud$ is a stochastic point cloud such that, for each point:
	\begin{enumerate*}
	\item its location is a 3D Gaussian random variable $\pcrandom_m \sim \Normal\paren{\pc_m, \varepsilon I}$, where $I$ is the $3\times 3$ identity matrix;
	\item its normal $\pcrandomnormal_m$ is a spherical random variable with conditional mean direction $\pcnormal_m$ and mean resultant length $\impfeat_m$, i.e., $\E{\conditional{\pcrandomnormal_m}{\pcrandom_m}} = \impfeat_m \pcnormal_m$.
	\end{enumerate*}
	We also assume that all other point cloud attributes are deterministic. Then, the expected value of $\pcwn$ in \cref{eqn:pcwn} equals:
	\begin{equation}\label{eqn:expected_pcwn}
		\Exp{\curly{\pcrandom_m,\pcrandomnormal_m}_{m=1}^M}{\pcwn\paren{\point}} = \sum_{m=1}^M \pcarea_m \poisson_\varepsilon\paren{\point, \pc_m}\impfeat_m = \regpcdl{\impfeat}.
	\end{equation}
\end{prp}

We refer to \citet[Chapter 9]{mardia2009directional} for background on spherical random variables, and prove \cref{pro:expected_winding_number} in \cref{app:noise}. \Cref{pro:expected_winding_number} lends theoretical support to our use of a regularized dipole sum to represent the geometry of point clouds $\pcloud$ with noisy ($\varepsilon > 0$) and outlier ($\impfeat_m \approx 0$)%
\footnote{In this stochastic interpretation, $\impfeat_m = 0$ means that the random normal $\pcrandomnormal_m$ (conditional on $\pcrandom_m$) is \emph{uniformly} distributed on the sphere. Then, the direction of the corresponding dipole is completely uncertain, and on expectation the dipole vanishes.}
 points. It also suggests the option to use different values $\varepsilon_m$, or even covariance matrices $\Sigma_m$, to model varying and anisotropic per-point uncertainty \citep{fuhrmann2014floating}. Empirically, we did not find doing so beneficial, and thus use a global $\varepsilon$ value that we select as we explain in \cref{sec:inverse}.

\section{Inverse rendering with point-based fields}\label{sec:representation}

We derived a point-based representation for the geometry field in inverse rendering. To complete our inverse rendering pipeline, we introduce a point-based representation for the radiance field (\cref{sec:radiance}), and explain how to optimize both fields (\cref{sec:inverse}).

\subsection{Radiance field representation}\label{sec:radiance}

Our derivation of the regularized dipole sum in \cref{sec:regularized} focused on point-based representation of scene geometry. However, \cref{eqn:dipole_sum} provides a way to interpolate any \emph{learnable} per-point attributes---corresponding to point samples of the continuous moment of a double-layer potential, or equivalently the jump-Dirichlet boundary condition of a Laplace equation---to scalar fields for use in inverse rendering. Thus, the regularized dipole sum lends itself as a point-based representation also for the radiance field.

To this end, we first interpolate a set of per-point \emph{appearance attributes} $\radfeat^k$ using regularized dipole sums $\regpcdl{\radfeat}^k\paren{\point}$ as in \cref{eqn:dipole_sum},
\begin{equation}\label{eqn:appearance_features}
	\regpcdl{\radfeat}^k\paren{\point} \equiv \sum_{m=1}^M \pcarea_m \poisson_\varepsilon\paren{\point, \pc_m} \radfeat^k_m, \; k=1, \dots, K.
\end{equation}
We then represent the radiance field $\radiance\paren{\point, \direction}$ as the output of a shallow multi-layer perceptron (MLP) that takes as input the values $\regpcdl{\radfeat}^k\paren{\point}$, position $\point$ and (encoded) direction $\direction$, and the \emph{implicit surface normal} from the geometry field $\impnormal\paren{\point} \equiv \nicefrac{\nabla \impfunc\paren{\point}}{\norm{\nabla \impfunc\paren{\point}}}$:%
\footnote{We experimented with a representation where the interpolated appearance attributes $\regpcdl{\radfeat}^k\paren{\point}$ are \emph{spherical harmonic coefficients} that are convertible to radiance $\radiance\paren{\point,\direction}$ through a rotation operation, as advocated by \citet{karnewar2022relu} and \citet{fridovich2022plenoxels}. Unfortunately, this approach, though sufficient for rendering high-quality novel views, resulted in surface artifacts around regions of strong specular appearance. \citet[Section 5]{Dai2024GaussianSurfels} report similar issues, which they alleviate by using monocular normal priors. We instead followed \citet{wu2022voxurf} and used an MLP to post-process the interpolated appearance attributes, to elide supervised data-driven priors.}
\begin{equation}\label{eqn:radiance_field}
	\radiance\paren{\point, \direction} \equiv \mlp\paren{\point, \direction, \impnormal\paren{\point}, \regpcdl{\radfeat}^1\paren{\point}, \dots, \regpcdl{\radfeat}^K\paren{\point}}.
\end{equation}
The radiance field $\radiance$ and the geometry field $\impfunc$ (\cref{eqn:geometry_field}) are intertwined, as the regularized dipole sums for appearance (\cref{eqn:appearance_features}) and geometry (\cref{eqn:dipole_sum}) share the same weights---determined by point cloud locations $\pc_m$, area weights $\pcarea_m$, and normals $\pcnormal_m$.

\paragraph{Relationship to mean value interpolation} Our use of regularized dipole sums in \cref{eqn:appearance_features} to interpolate per-point appearance attributes $\radfeat$ is closely related to 3D interpolation with \emph{mean value coordinates} \citep{floater2005mean,ju2005mean}. Using \cref{eqn:wn,eqn:double_layer_potential}, we can write the mean value interpolant (and its point-cloud approximation) at $\point \in \R^3$ of a function $\radfeat : \boundary \to \R$ as:
\begin{equation}\label{eqn:mvc}
	\mv^{\radfeat}\paren{\point} \equiv \frac{\dl{\radfeat}\paren{\point}}{\wn\paren{\point}} \approx \frac{\pcdl{\radfeat}\paren{\point}}{\pcwn\paren{\point}}.
\end{equation}
Mean value interpolation has found widespread use in computer graphics and other areas \citep{hormann2017generalized,chen2024fast,deGoes2024barycentric}, a success in large part thanks to the geometric regularization properties of the mean value interpolant \citep[Section 2]{ju2005mean}. These properties and empirical success lend support to our choice of (regularized) dipole sums as a point-based representation for the radiance field. In our representation, we omit normalization (denominator in \cref{eqn:mvc}), as we found empirically that the \emph{linear precision} property it enforces inhibits the ability of the radiance field to reproduce specular highlights.

\begin{algorithm*}[t]
	\caption{Barnes-Hut accelerated primal and adjoint queries for fast dipole sums.}
	\label{alg:queries}
	\begin{algorithmic}[1]
	\algblockdefx[Name]{Struct}{EndStruct}
		[1][Unknown]{\textbf{struct} #1}
		{}
	\algtext*{EndStruct}
	\algblockdefx[Name]{FORDO}{ENDFORDO}
		[1][Unknown]{\textbf{for} #1 \textbf{do}}
		{}
	\algtext*{ENDFORDO}
	\algblockdefx[Name]{PARALLELFORDO}{ENDPARALLELFORDO}
		[1][Unknown]{\textbf{parallel}\ \textbf{for} #1 \textbf{do}}
		{}
	\algtext*{ENDPARALLELFORDO}
	\algblockdefx[Name]{IF}{ENDIF}
		[1][Unknown]{\textbf{if} #1 \textbf{then}}
		{}
	\algtext*{ENDIF}
	\algblockdefx[Name]{IFTHEN}{ENDIFTHEN}
		[2][Unknown]{\textbf{if} #1 \textbf{then} #2}
		{}
	\algtext*{ENDIFTHEN}
	\algblockdefx[Name]{RETURN}{ENDRETURN}
		[1][Unknown]{\textbf{return} #1}
		{}
	\algtext*{ENDRETURN}
	\algblockdefx[Name]{COMMENT}{ENDCOMMENT}
		[1][Unknown]{\textcolor{commentcolor}{\(\triangleright\)\textit{#1}}}
		{}
	\algtext*{ENDCOMMENT}

	\Struct[$\Proc{TreeNode}$]
		\State $\nodepc, \nodearea, \noder, \nodedirichlet \gets \Proc{TreeUpdate}$\Comment{Immutable node attributes initialized using \cref{eqn:farfield_dipole,eqn:farfield_dipole_features}}
		\State $\dd\nodedirichlet \gets 0$ \Comment{Mutable node gradient attribute}
		\Function{GetContribution}{$\point,\varepsilon$}
			\RETURN[$\nodearea \regul\paren{\nicefrac{\norm{\nodepc-\point}}{\varepsilon}} \nicefrac{\paren{\nodepc-\point}}{\norm{\nodepc-\point}^3} \cdot \nodedirichlet$]\Comment{Compute node contribution to dipole sum using \cref{eqn:farfield}}
			\ENDRETURN
		\EndFunction
		\Function{IncrementGradient}{$\dd\regpcdld, \point,\varepsilon$}
			\State $\dd\nodedirichlet \mathrel{+}= \nodearea \regul\paren{\nicefrac{\norm{\nodepc-\point}}{\varepsilon}} \nicefrac{\paren{\nodepc-\point}}{\norm{\nodepc-\point}^3} \cdot \dd\regpcdld$\Comment{Increment node gradient attribute using \cref{eqn:farfield_gradient}}
		\EndFunction
		\Function{GetChildren}{}
			\RETURN[\texttt{listOfChildrenNodes}]\Comment{Return a list of children nodes, or empty list if node is a leaf}
			\ENDRETURN
		\EndFunction
	\EndStruct

 	\Require \textit{A query point $\point$, the root node of a tree structure \texttt{node}, a control parameter $\beta$.}
 	\Ensure \textit{Dipole sum $\regpcdld\paren{\point}$.}
	\Function{PrimalQuery}{$\point,\texttt{node},\varepsilon, \beta$}
		\IFTHEN[$\norm{\point - \texttt{node}.\nodepc} > \beta \cdot \texttt{node}.\noder$]{\textbf{return} $\texttt{node}.\Proc{GetContribution}(\point,\varepsilon)$\Comment{If the query point is far from the cluster, terminate}}
		\ENDIFTHEN
		\State $\texttt{listOfChildrenNodes} \gets \texttt{node}.\Proc{GetChildren}$ \Comment{Get list of children nodes}
		\IFTHEN[$\Proc{IsEmpty}(\texttt{listOfChildrenNodes})$]{\textbf{return} $\texttt{node}.\Proc{GetContribution}(\point,\varepsilon)$\Comment{If the node is a leaf, terminate}}
		\ENDIFTHEN
		\State $\regpcdld \gets 0$ \Comment{Initialize dipole sum value}
		\FORDO[$\texttt{child}\ \textbf{in}\ \texttt{listOfChildrenNodes}$]
			\State $\regpcdld \mathrel{+}= \Proc{PrimalQuery}(\point,\texttt{child},\varepsilon,\beta)$ \Comment{Iterate over all children nodes}
		\ENDFORDO
		\RETURN[$\regpcdld$]\ENDRETURN
	\EndFunction

	\Require \textit{A gradient $\dd\regpcdld$, a query point $\point$, the root node of a tree structure \texttt{node}, a control parameter $\beta$.}
 	\Function{AdjointQuery}{$\dd\regpcdld,\point,\texttt{node},\varepsilon, \beta$}
		\IFTHEN[$\norm{\point - \texttt{node}.\nodepc} > \beta \cdot \texttt{node}.\noder$]{$\texttt{node}.\Proc{IncrementGradient}(\dd\regpcdld,\point,\varepsilon)$ \textbf{return}\Comment{If the query point is far from the cluster, terminate}}
		\ENDIFTHEN
		\State $\texttt{listOfChildrenNodes} \gets \texttt{node}.\Proc{GetChildren}$ \Comment{Get list of children nodes}
		\IFTHEN[$\Proc{IsEmpty}(\texttt{listOfChildrenNodes})$]{$\texttt{node}.\Proc{IncrementGradient}(\dd\regpcdld,\point,\varepsilon)$ \textbf{return}\Comment{If the node is a leaf, terminate}}
		\ENDIFTHEN
		\FORDO[$\texttt{child}\ \textbf{in}\ \texttt{listOfChildrenNodes}$]
			\State $\Proc{AdjointQuery}(\dd\regpcdld,\point,\texttt{child},\varepsilon,\beta)$ \Comment{Iterate over all children nodes}
		\ENDFORDO
	\EndFunction
	\end{algorithmic}
\end{algorithm*}

\subsection{Inverse rendering optimization}\label{sec:inverse}

Our overall scene representation comprises a point cloud $\pcloud \equiv \curly{\paren{\pc_m,\pcnormal_m,\pcarea_m, \impfeat_m, \radfeat^1_m,\dots, \radfeat^K_m}}_{m=1}^M$ with per-point locations, normals, area weights, geometry attribute, and appearance attributes; as well as the parameters of the MLP in \cref{eqn:radiance_field}. We use this representation to compute the geometry field $\impfunc$ (\cref{eqn:dipole_sum,eqn:geometry_field}) and radiance field $\radiance$ (\cref{eqn:appearance_features,eqn:radiance_field}). During the inverse rendering stage, we synthesize images using volume rendering and ray tracing (\cref{eqn:quadrature}) combined with $\impfunc$ and $\radiance$. We then optimize the point cloud $\pcloud$ and MLP by minimizing the loss:
\begin{equation}\label{eqn:loss}
	\loss_{\mathrm{rendering}} + \loss_{\mathrm{entropy}} + \loss_{\mathrm{winding}} + \loss_{\mathrm{normal}},
\end{equation}
where each summand includes an appropriate weight, and:
\begin{enumerate}[leftmargin=*]
	\item $\loss_{\mathrm{rendering}}$ is the $L^1$-loss between input and rendered images;
	\item $\loss_{\mathrm{entropy}}$ is a per-ray entropy loss inspired from \citet{kim2022infonerf} (we provide details in \cref{app:entropy});
	\item $\loss_{\mathrm{winding}}$ aggregates losses $\norm{\impfeat_m - 1}^2$ on the point cloud;
	\item $\loss_{\mathrm{normal}}$ aggregates losses $\norm{\pcnormal_m - \pcnormal_{m,\mathrm{init}}}^2$ on the point cloud.
\end{enumerate}
The loss $\loss_{\mathrm{winding}}$ regularizes the geometry field $\impfunc$ by penalizing large deviations between the regularized dipole sum $\regpcdl{\impfeat}$ and the regularized winding number $\regpcwn$. The loss $\loss_{\mathrm{normal}}$ penalizes large changes to point normals compared to their initial values.

We initialize $\pcloud$ with locations $\pc_m$ and normals $\pcnormal_m$ using the \emph{dense} point cloud output of COLMAP~\citep{schoenberger2016sfm}, and area weights $\pcarea_m$ computed as in \citet{barill2018fast}. We initialize the geometry attributes $\impfeat_m$ to 1 (equal to the regularized winding number), and appearance attributes $\radfeat^k_m$ using Gaussian random variates. Inverse rendering optimizes: the normals, geometry attributes, and appearance attributes of $\pcloud$; the global scale $\scale$ and regularization $\varepsilon$ parameters in \cref{eqn:vacancy,eqn:regularized_poisson} (resp.); and the MLP parameters in \cref{eqn:radiance_field}. Importantly, we do not optimize the area weights and locations in $\pcloud$, to facilitate fast inverse rendering---we elaborate in \cref{sec:backprop}. Instead, the geometry and appearance attributes provide us with enough degrees of freedom to represent high-quality geometry and appearance, and correct defects (noisy points, outliers, holes) in the dense structure-from-motion point cloud.

\section{Barnes-Hut fast summation}\label{sec:backprop}

Rendering with our point-based representations requires \emph{evaluating} dipole sums (\cref{eqn:dipole_sum,eqn:appearance_features}) at multiple locations along each viewing ray, to compute the geometry (\cref{eqn:geometry_field}) and radiance (\cref{eqn:radiance_field}) fields in \cref{eqn:rendering}. Inverse rendering with these representations requires additionally \emph{backpropagating} through each dipole sum, to compute derivatives of per-point attributes. We term such evaluation and backpropagation operations \emph{primal} and \emph{adjoint} (resp.) \emph{dipole sum queries}, using terminology from differentiable rendering \citep{nimier2020radiative,vicini2021path,stam2020computing}. Implemented naively (i.e., as summations by iterating over all points), primal and adjoint queries have linear complexity $\bigO\paren{M}$ relative to point cloud size $M$. Consequently, during inverse rendering, these queries become the main computational burden when working with even moderately large point clouds; and become prohibitively expensive when working with the dense point clouds output by structure-from-motion initialization.

Fortunately, it is possible to dramatically accelerate both types of queries, enabling inverse rendering at speeds competitive with rasterization methods \citep{Dai2024GaussianSurfels}. In particular, \citet{barill2018fast} show how to perform primal queries for the winding number with \emph{logarithmic complexity} $\bigO\paren{\log M}$, using the classical \emph{Barnes-Hut fast summation method} \citep{barnes1986hierarchical}. We adopt their approach, which we adapt below to regularized dipole sums. Then, we show how to use Barnes-Hut fast summation to perform also adjoint queries with logarithmic complexity. To simplify discussion, throughout this section we use $\dirichlet$ as a stand-in for any of the \emph{moment attributes} stored in $\pcloud$---namely, the geometry attribute $\impfeat$ and the appearance attributes $\radfeat^k,\ k=1,\dots,K$.

\subsection{Acceleration of primal queries}\label{sec:primal}

The Barnes-Hut method first creates a tree data structure (e.g., octree \citep{meagher1982geometric}) whose nodes hierarchically subdivide the point cloud $\pcloud$ into clusters, with leaf nodes corresponding to individual points. Each tree node $t$ is assigned a centroidal radius and attributes representative of the set $\LS\paren{t}$ of all leaf nodes that are successors of $t$ in the tree hierarchy. We follow \citet{barill2018fast} and assign the node area, location, area, and radius (resp.) attributes:
\begin{equation}\label{eqn:farfield_dipole}
	\nodearea_t \equiv \!\!\!\sum_{m \in \LS\paren{t}} \!\!\!\pcarea_m,\ \nodepc_t \equiv \frac{1}{\nodearea_t}\!\!\sum_{m \in \LS\paren{t}}\!\!\!\pcarea_m \pc_m, \ \noder_t \equiv \!\!\!\max_{m \in \LS\paren{t}} \norm{\pc_m - \nodepc_t},
\end{equation}
as well as \emph{vector-valued} moment attributes:
\begin{equation}\label{eqn:farfield_dipole_features}
	\nodedirichlet_t \equiv \frac{1}{\nodearea_t}\sum_{m \in \LS\paren{t}} \pcarea_m \pcnormal_m \dirichlet_m,
\end{equation}
%
which absorb the leaf nodes' moment \emph{and} normal attributes.

Then, for a primal query at point $\point$, the Barnes-Hut method performs a depth-first tree traversal: at each node $t$, if $\point$ is sufficiently far from the node's centroid (i.e., $\norm{\point - \nodepc_t} > \beta\noder_t$, where $\beta$ is a user-defined parameter), the node's successors are not visited. Instead, the sum of contributions from all leaf nodes in $\LS\paren{t}$ to the dipole sum is approximated using the node's attributes:
\begin{equation}\label{eqn:farfield}
	\sum_{m \in \LS\paren{t}} \pcarea_m \poisson_{\varepsilon}\paren{\point,\pc_m} \dirichlet_m  \approx \nodearea_t \regul\paren{\frac{\norm{\nodepc_t-\point}}{\varepsilon}} \frac{\nodedirichlet_t\cdot\paren{\nodepc_t-\point}}{\norm{\nodepc_t-\point}^3}.
\end{equation}
%
This approximation expresses the fact that, due to the squared-distance falloff of $\poisson_\varepsilon$, the \emph{far-field} influence of a cluster of points can be represented by a single point at the cluster's centroid. \Cref{alg:queries} (lines 13--21) summarizes the accelerated primal queries.

\subsection{Acceleration of adjoint queries}

A naive implementation of adjoint queries by using automatic differentiation (e.g., autograd \citep{paszke2017automatic}) would result in linear complexity $\bigO\paren{M}$, even though the differentiated primal query has logarithmic complexity $\bigO\paren{\log M}$. The reason is that, even if the primal query stopped tree traversal at a node $t$, the node's attributes would be functions of those of all successor leaf nodes. Thus, the adjoint query would still end up visiting all leaf nodes.

To maintain logarithmic complexity during inverse rendering, we use at each gradient iteration a two-stage backpropagation scheme:
\begin{enumerate}[leftmargin=*]
	\item At the start of the iteration, after the tree updates, we \emph{detach} the node attributes from the corresponding leaf node attributes. During inverse rendering, each adjoint query backpropagates gradients to \emph{only} the nodes visited by the corresponding primal query. Each node locally accumulates rendering gradients.
	\item After inverse rendering concludes, we perform a \emph{single} full tree traversal to propagate accumulated gradients from all nodes to the leaf nodes. The resulting gradients are used to update point cloud attributes at the end of the iteration.
\end{enumerate}
This two-stage process requires storing at each tree node a set of additional \emph{mutable} gradient attributes $\dd\nodedirichlet_t$ (one for each of the geometry and appearance attributes),
to accumulate backpropagated gradients. Overall, if we perform a total of $Q$ queries during each gradient iteration, naive backpropagation would result in complexity $\bigO\paren{Q M}$. Our two-stage backpropagation has instead complexity $\bigO\paren{Q \log M + M\log M}$: $\bigO\paren{Q \log M}$ for the adjoint queries in the first stage; and $\bigO\paren{M \log M}$ for the full traversal at the second stage (updating $M$ leaf nodes, each with $\bigO\paren{\log M}$ ancestors).

\Cref{alg:queries} (lines 22--28) summarizes the accelerated adjoint queries in the first stage, which we implement exactly analogously to primal queries. The second stage is likewise easy to implement automatic differentiation. We provide details in \cref{app:backprop}.

\subsection{Acceleration details}\label{sec:backprop_details}

We conclude this section by highlighting some salient details regarding our Barnes-Hut acceleration scheme.

\paragraph{Tree construction and update} We construct the octree data structure after structure from motion, using its dense point cloud output. The node hierarchy in the tree depends on only the point cloud locations $\pc_m$. Thus, as we choose not to update these locations during inverse rendering (\cref{sec:inverse}), we create the tree structure only \emph{once} rather than after each gradient operation. Choosing otherwise would introduce significant computational overhead.

At each iteration, we must twice update the moment attributes $\nodedirichlet_t$ of the tree nodes: once at the start of the iteration, to account for updated point cloud attributes after a gradient step; and once at its end, during the second-stage of backpropagation. Both updates are efficient, introducing an overhead analogous to about a couple additional ray casting queries during rendering.

\paragraph{Queries for multiple moment attributes} Primal and inverse rendering with \cref{eqn:rendering} requires performing, at every sampled ray location $\point$, dipole sum queries for all moment attributes $\dirichlet$ stored in the point cloud---namely, the geometry attribute $\impfeat$ and the appearance attributes $\radfeat^k,\ k=1,\dots,K$. As the tree traversal pattern depends on only $\point$, we can return all these attributes with a single primal query and tree traversal---and likewise for adjoint queries. Further accelerating performance by using packet queries \citep{wald2014embree} for multiple query points $\point$ is an exciting future direction.

\section{Experimental evaluation}\label{sec:experiments}

We evaluate our method against state-of-the-art methods for multi-view surface reconstruction: Gaussian surfels \citep{Dai2024GaussianSurfels}, which combines a point-based representation with rasterization; and Neus2 \citep{neus2} and Neuralangelo \citep{li2023neuralangelo}, which both combine a hybrid hashgrid-neural representation with ray tracing. All three methods aim for high-quality surface outputs, but place different emphasis on computational efficiency (Gaussian surfels, NeuS2) versus reconstruction fidelity (Neuralangelo). In summary, our results suggest that our method provides, at equal runtimes, improved reconstruction quality and robustness compared to these alternatives. Additionally, though we do not include direct comparisons, our results (\cref{tab:dtu}) additionally suggest that our method improves reconstruction quality compared to other concurrent 3D Gaussian methods, including 2D Gaussian splatting \citep[Table 1]{huang20242dgs} and Gaussian opacity fields \citep[Table 2]{yu2024gaussian}. We provide additional results and code on the project website.

\subsection{Implementation details}\label{sec:implementation}

We built our codebase in PyTorch \citep{paszke2019pytorch} based on the NeuS codebase \citep{NeuSCodebase}. We implemented custom C++ and CUDA extensions for building the octree and performing fast primal and adjoint dipole sum queries, following the original C++ implementation of fast winding numbers \citep{barill2018fast} in libigl \citep{libigl}. Our code is available on the project website.

\paragraph{Radiance field details} We design the MLP in \cref{eqn:radiance_field} similarly to the appearance network of NeuS \citep{wang2021neus}---4 hidden layers, each with 256 neurons and ReLU activations. We encode viewing directions with real spherical harmonics up to degree 3, and use $K=32$ appearance attributes $\radfeat^k$, for which we found it beneficial to skip the foreshortening term when computing dipole sums (\cref{eqn:appearance_features}). We apply weight normalization \citep{salimans2016weightNormalization} for stable training. We limit the radiance field inside a bounding sphere, and use a background network based on NeRF++ \citep{kaizhang2020nerfpp} to model the exterior of the sphere.

\paragraph{Ray sampling} We sample ray locations in \cref{eqn:quadrature} as in \citet{miller2023theory}: For each camera ray, we identify the first zero-crossing of the geometry field $\impfunc$ by densely placing 1024 samples along the ray between the near and far limits. If a zero-crossing is found, we place 24 sparse samples between the near limit and the first crossing, 48 dense samples around the first crossing, and 8 sparse samples between the first crossing and the far limit. Otherwise, we place 80 samples uniformly between the near and far limits.

An advantage of representing the geometry field as a dipole sum is that we can compute the first zero-crossing along a ray efficiently (with logarithmic complexity in terms of number of dipole sum queries) using \emph{Harnack tracing} \citep[Section 4.3]{Gillespie:2024:RTH}---a method analogous to sphere tracing for signed distance functions \citep{hart1996sphere}, except designed for (near-)harmonic functions. In practice, because our fast primal queries contribute only minor overhead to the overall inverse rendering runtime, we found that Harnack tracing provided negligible acceleration compared to the ray marching procedure we described above; thus we use the latter for simplicity.

\begin{figure}
	\centering
	\includegraphics[width=\linewidth]{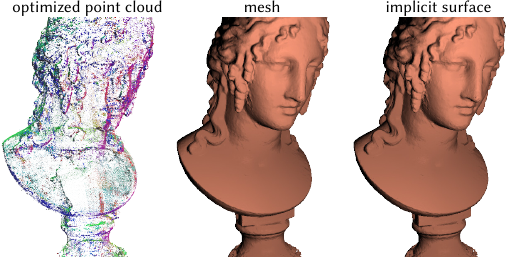}
	\caption{Our regularized dipole sum representation allows us to directly ray trace the optimized point cloud (where we use color to visualize normals, and size to visualize geometry attributes), achieving the same results as ray tracing a mesh without the need to extract one.}
	\label{fig:tracing}
	\vspace{-1em}
\end{figure}

\begin{figure*}
	\centering
	\includegraphics[width=\textwidth]{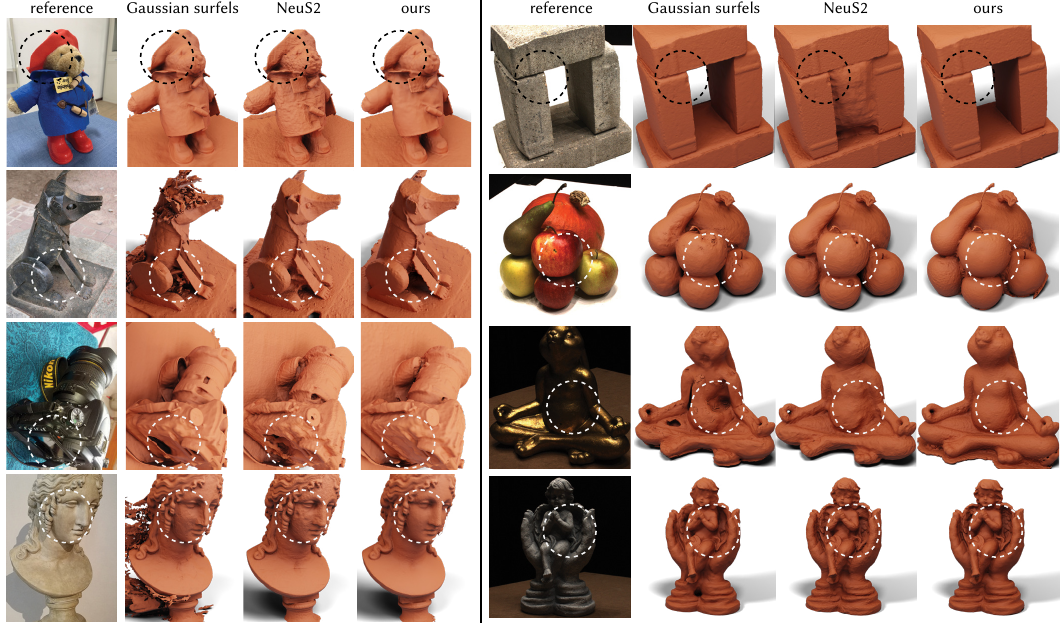}
	\caption{Qualitative comparisons on the BlendedMVS (left) and DTU (right) datasets. The dashed circles indicate areas of interest. NeuS2 captures fine details, but produces noisy meshes with structural artifacts. Gaussian surfels produces floater artifacts that require manual filtering. By contrast, our method produces clean meshes with correct and artifact-free geometry. We provide interactive visualizations of results on the entire datasets on the project website.}
	\label{fig:qualitative}
	\vspace{-1em}
\end{figure*}

\paragraph{Training} We use Adam \citep{kingma2015adam} with a batch size 4096 rays for optimization. We use a learning rate of \(1\times 10^{-2}\) for point cloud attributes, and \(3\times 10^{-3}\) for the radiance field MLP. We use a linear warmup schedule for the first 200 iterations, and a cosine decay schedule for the remaining iterations. We use different numbers of iterations depending on the experiment---training for 1000, 3000, and 20000 iterations takes 3 min, 8 min, and 1 hour (resp.) on a single NVIDIA RTX 4090 GPU.

\paragraph{Point growing} As we mentioned in \cref{sec:moment}, the use of non-unit geometry attributes for the geometry field $\impfunc$ helps fill point cloud holes due to textureless regions. In practice, we found it useful to \emph{also} grow a small number of additional points during inverse rendering. We perform point growing every 500 iterations, by sampling random rays and computing their first intersection with the geometry field. At each intersection, we add a point if
%
the distance to the closest point in the point cloud is greater than a threshold.
%
For each new point, we initialize its attributes by averaging those of its neighbors, compute a normal using PCA \citep{hoppe1992surface}, then recompute the area weights of the entire point cloud. We found that we need to grow only about 10\% additional points relative to the original dense point cloud from structure from motion, as we show in \cref{fig:growing}.

\paragraph{Mesh extraction} We produce meshes by extracting the zero-level set of $\impfunc$ using marching cubes \citep{lorensenMarchingCubes}, at a grid resolution of \(512^3\) for DTU and \(1024^3\) for BlendedMVS. We make two observations:
\begin{enumerate*}
	\item \Citet{barill2018fast} suggest using bisection root-finding to extract meshes from the winding number field, to avoid artifacts due to the singular Poisson kernel. By contrast, thanks to the regularized Poisson kernel, we can extract artifact-free meshes using marching cubes, as we show in \cref{fig:regularized}.
	\item We need to extract meshes \emph{only} for quantitative comparisons with other methods. We can directly and efficiently ray trace our geometry field using ray marching with fast primal queries (and optionally Harnack tracing \citep{Gillespie:2024:RTH}), as we show in \cref{fig:tracing}.
\end{enumerate*}

\subsection{Comparison to prior work}\label{sec:comparison}

\begin{figure}
	\centering
	\includegraphics[width=\linewidth]{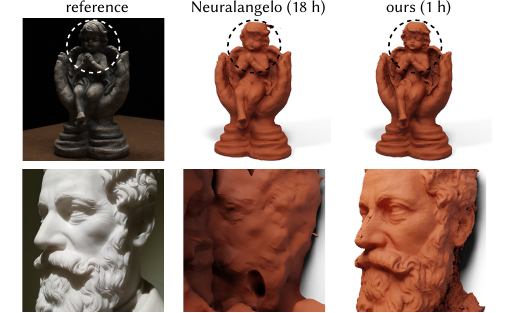}
	\caption{Our method produces higher-quality reconstructions than Neuralangelo on DTU scenes at \(\nicefrac{1}{18}\) of the runtime (top row). Neuralangelo fails on BlendedMVS scenes when few views are available (bottom row).}
	\label{fig:neuralangelo}
	\vspace{-1em}
\end{figure}

\begin{figure}
	\centering
	\includegraphics[width=\linewidth]{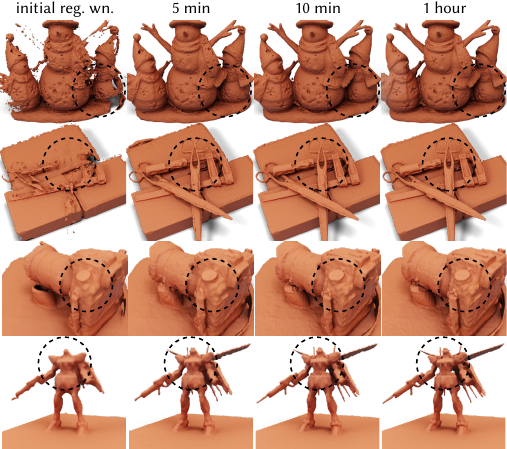}
	\caption{Progression of our method on DTU (top two rows) and BlendedMVS (bottom two rows) scenes. The leftmost column shows the mesh extracted from the initial regularized winding number field, and the remaining columns show meshes at runtimes of 5 minutes, 10 minutes, and 1 hour. Our method significantly improves the initial mesh within 5 minutes (3 minutes training), and continues to refine details with additional training.}
	\label{fig:progression}
	\vspace{-1em}
\end{figure}

We evaluate our method against NeuS2 \citep{neus2}, Gaussian surfels \citep{Dai2024GaussianSurfels}, and Neuralangelo \citep{li2023neuralangelo}, on the DTU \citep{DTU} and BlendedMVS \citep{BlendedMVS} datasets. 
We train our method, NeuS2, and Gaussian surfels without mask supervision and evaluate their extracted meshes using the DTU evaluation script. For each method, we present results for \emph{runtimes} of 5 minutes, 10 minutes, and 1 hour, which we measure as follows to ensure fair comparisons: For NeuS2 and Gaussian surfels, runtime equals training time; for our method, runtime includes the time of the refinement process in COLMAP (which the other two methods do not require), thus decreasing training time (e.g., about 2 minutes COLMAP refinement and 3 minutes training for a total runtime of 5 minutes). For Neuralangelo, we report DTU evaluation scores from their paper, and do not report scores on BlendedMVS as on multiple scenes it failed to produce meaningful reconstructions (e.g., second row of \cref{fig:neuralangelo}). Lastly, we also compare against meshes extracted using our regularized winding number (\cref{eqn:regularized_pcwn}) on the dense point cloud output by COLMAP without training---effectively, the initialization of our method. We provide quantitative results in \cref{tab:dtu,tab:blendedmvs}, qualitative results in \cref{fig:qualitative,fig:neuralangelo,fig:progression}, and interactive visualizations on the project website.

Quantitatively, we observe that our method overall outperforms Gaussian surfels and NeuS2 at all runtimes on both DTU and BlendedMVS. Moreover, our method consistently improves reconstruction quality with additional training time. By contrast, NeuS2 and Gaussian surfels either stagnate or even degrade performance with additional training time. Our method at 1 hour of runtime also outperforms Neuralangelo at 18 hours of training on the DTU dataset.

In all cases, the quantitative improvements also translate to visual qualitative improvements on the extracted meshes. \Cref{fig:qualitative,fig:neuralangelo} show some examples, but we encourage using the interactive visualization on the project website to better assess qualitative differences. Our method occasionally takes longer to recover finer details, because it keeps point cloud positions fixed to regularize geometry and prevent introduction of geometric defects. Even with this regularization, as training time increases, our method only improves reconstructed geometry; and with sufficient training time, it reliably recovers finer geometric details. \Cref{fig:progression} shows some examples visualizing the training progression of our method.

\begin{table}[t]
	\centering
	\caption{Chamfer distances on DTU for different runtimes. (N.2: Neus2, G.S.: Gaussian surfels, N.A.: Neuralangelo, init.: regularized winding number on the dense COLMAP point cloud without training.)}
	\vspace{-0.75em}
	\begin{tabularx}{\linewidth}{@{} X |  >{\centering\arraybackslash}X | >{\centering\arraybackslash}X >{\centering\arraybackslash}X >{\centering\arraybackslash}X | >{\centering\arraybackslash}X >{\centering\arraybackslash}X >{\centering\arraybackslash}X | >{\centering\arraybackslash}X >{\centering\arraybackslash}X @{}}
		\toprule
		& init. & \multicolumn{3}{c|}{\qty{5}{m}} & \multicolumn{3}{c|}{\qty{10}{m}} & \qty{18}{h} & \qty{1}{h}\\
		& \!\textbf{ours} & N.2 & G.S. & \!\textbf{ours} & N.2 & G.S. & \!\textbf{ours} & N.A. & \!\textbf{ours} \\
		\midrule
		24 & 1.82 & 0.78 & 0.68 & 0.80 & 0.75 & 0.62 & 0.64 & 0.37 & 0.45 \\
		37 & 1.34 & 0.64 & 0.77 & 0.77 & 0.65 & 0.76 & 0.69 & 0.72 & 0.67 \\
		40 & 0.54 & 1.04 & 0.56 & 0.38 & 1.06 & 0.49 & 0.35 & 0.35 & 0.32 \\
		55 & 0.60 & 0.30 & 0.47 & 0.39 & 0.28 & 0.48 & 0.36 & 0.35 & 0.31 \\
		63 & 0.76 & 1.01 & 0.86 & 0.91 & 1.00 & 0.84 & 0.90 & 0.87 & 0.93 \\
		65 & 1.37 & 0.62 & 1.06 & 0.94 & 0.59 & 1.08 & 0.79 & 0.54 & 0.67 \\
		69 & 1.45 & 0.68 & 0.86 & 0.78 & 0.67 & 0.88 & 0.76 & 0.53 & 0.53 \\
		83 & 0.95 & 1.17 & 1.09 & 0.69 & 1.18 & 1.09 & 0.72 & 1.29 & 0.79 \\
		97 & 1.78 & 1.00 & 1.31 & 1.00 & 1.04 & 1.31 & 0.97 & 0.97 & 0.91 \\
		105 & 0.88 & 0.71 & 0.74 & 0.61 & 0.74 & 0.75 & 0.59 & 0.73 & 0.63 \\
		106 & 0.80 & 0.55 & 0.83 & 0.73 & 0.54 & 1.05 & 0.60 & 0.47 & 0.48 \\
		110 & 1.43 & 0.89 & 1.76 & 0.93 & 0.84 & 1.76 & 0.83 & 0.74 & 0.57 \\
		114 & 0.60 & 0.36 & 0.52 & 0.47 & 0.37 & 0.52 & 0.39 & 0.32 & 0.32 \\
		118 & 0.94 & 0.47 & 0.64 & 0.55 & 0.43 & 0.67 & 0.49 & 0.41 & 0.40 \\
		122 & 0.71 & 0.45 & 0.59 & 0.49 & 0.43 & 0.61 & 0.42 & 0.43 & 0.39 \\
		\midrule
		avg. & \textbf{1.06} & 0.71 & 0.85 & \textbf{0.70} & 0.70 & 0.86 & \textbf{0.63} & 0.61 & \textbf{0.56} \\
		\bottomrule
	\end{tabularx}
	\label{tab:dtu}
	\vspace{-1em}
\end{table}
\begin{table}[t]
	\centering
	\caption{Chamfer distances on BlendedMVS for different runtimes. (N.2: Neus2, G.S.: Gaussian surfels, init.: regularized winding number on the dense COLMAP point cloud without training; \xmark~indicates failure to converge.)}
	\vspace{-0.75em}
	\begin{tabularx}{\linewidth}{@{} X | >{\centering\arraybackslash}X | >{\centering\arraybackslash}X >{\centering\arraybackslash}X >{\centering\arraybackslash}X | >{\centering\arraybackslash}X >{\centering\arraybackslash}X >{\centering\arraybackslash}X | >{\centering\arraybackslash}X >{\centering\arraybackslash}X >{\centering\arraybackslash}X @{}}
		\toprule
		& init. & \multicolumn{3}{c|}{\qty{5}{m}} & \multicolumn{3}{c|}{\qty{10}{m}} & \multicolumn{3}{c}{\qty{1}{h}} \\
		& \!\!\textbf{ours} & N.2 & G.S. & \!\!\textbf{ours} & N.2 & G.S. & \!\!\textbf{ours} & N.2 & G.S. & \!\!\textbf{ours} \\
		\midrule
		bas. & 0.52 & 0.76 & 0.55 & 0.62 & 0.72 & 0.54 & 0.61 & 0.75 & 0.47 & 0.45 \\
		bea. & 0.51 & 0.88 & 0.62 & 0.44 & 0.89 & 0.65 & 0.42 & 0.94 & 0.71 & 0.39 \\
		bre. & 1.06 & 0.72 & 0.44 & 0.45 & 0.77 & 0.48 & 0.27 & 0.58 & 0.68 & 0.22 \\
		cam. & 0.60 & 0.86 & 0.92 & 0.53 & 0.84 & 0.83 & 0.56 & 0.82 & 0.89 & 0.57 \\
		clo. & 0.82 & 1.40 & 1.71 & 0.68 & 1.33 & 1.14 & 0.66 & 1.41 & 1.44 & 0.65 \\
		cow & 0.52 & 0.66 & 1.95 & 0.56 & 0.64 & 2.01 & 0.54 & 0.64 & 2.69 & 0.56 \\
		dog & 0.98 & 1.22 & 1.53 & 0.77 & 1.23 & 1.54 & 0.69 & 1.21 & 1.71 & 0.61 \\
		dol. & 0.84 & 0.74 & 0.85 & 0.70 & 0.71 & 0.87 & 0.70 & 0.70 & 0.84 & 0.75 \\
		dra. & 1.86 & 0.96 & 2.46 & 0.83 & 0.91 & 1.75 & 0.66 & 0.97 & 1.58 & 0.50 \\
		dur. & 1.22 & \xmark & 1.43 & 1.04 & \xmark & 1.38 & 1.00 & \xmark & 1.47 & 0.98 \\
		fou. & 1.22 & 1.23 & 1.54 & 0.97 & 1.30 & 1.68 & 0.96 & 1.22 & 1.71 & 0.88 \\
		gun. & 0.61 & 0.34 & 0.84 & 0.37 & 0.35 & 0.45 & 0.36 & 0.41 & 0.55 & 0.32 \\
		hou. & 0.71 & 0.96 & 0.82 & 0.72 & 1.01 & 0.83 & 0.70 & 1.07 & 0.89 & 0.51 \\
		jad. & 2.15 & 1.64 & 1.59 & 1.83 & 1.54 & 2.10 & 1.80 & 1.63 & 1.67 & 1.66 \\
		man & 1.10 & 0.55 & 0.97 & 1.09 & 0.54 & 1.08 & 0.82 & 0.55 & 1.55 & 0.55 \\
		mon. & 0.67 & 0.39 & 0.73 & 0.42 & 0.35 & 0.87 & 0.41 & 0.35 & 1.43 & 0.36 \\
		scu. & 0.67 & 0.62 & 1.23 & 0.66 & 0.59 & 1.86 & 0.62 & 0.58 & 2.01 & 0.56 \\
		sto. & 0.79 & 0.92 & 0.64 & 0.64 & 0.78 & 0.68 & 0.63 & 0.79 & 0.70 & 0.53 \\
		\midrule
		avg. & \textbf{0.94} & 0.87 & 1.16 & \textbf{0.74} & 0.85 & 1.15 & \textbf{0.69} & 0.86 & 1.28 & \textbf{0.61} \\
		\bottomrule
	\end{tabularx}
	\label{tab:blendedmvs}
	\vspace{-1em}
\end{table}

By contrast, the alternative methods occasionally recover finer details earlier in training, but are prone to introducing geometric defects that cannot be resolved with additional training time (e.g., NeuS2 and Gaussian surfels reconstructions of the dog head and camera screen, in second and third row (resp.) of \cref{fig:qualitative}). These defects can even result in complete failure to extract a meaningful mesh (e.g., Neuralangelo in second row of \cref{fig:neuralangelo}). Additionally, the alternative methods often add higher frequency details not present in the input images (e.g., NeuS2 reconstruction of the bear first row of \cref{fig:qualitative}) giving the false impression of increased detail.

Lastly, our method reconstructs more accurate meshes than its initialization. Notably, this initialization already provides a high quality 3D reconstruction, and in several cases better than what NeuS2 and Gaussian surfels after an hour of training (\cref{tab:blendedmvs})! This behavior highlights the importance of leveraging \emph{dense} point cloud initialization from structure from motion in subsequent inverse rendering. To summarize, our method overall ensures robust performance by providing reliable geometry improvement and fine feature recovery, and outperforms alternative methods at equal (NeuS2, Gaussian surfels) or order-of-magnitude shorter (Neuralangelo) runtimes.


\subsection{Ablation study}\label{sec:ablation}

To evaluate the impact of different components of our method in overall performance, we perform an ablation study using the BlendedMVS dataset and the same experimental protocol as in \cref{sec:comparison}. We evaluate the following variants of our method:
\begin{enumerate*}
	\item removing each of the entropy, winding, and normal losses in \cref{eqn:loss} during inverse rendering optimization;
	\item removing kernel regularization;
	\item removing point growing; and
	\item removing normal training and keeping normals fixed to their initial values.
\end{enumerate*}
We provide quantitative results in \cref{tab:ablation}. We observe that performance deteriorates in all cases, suggesting that each of the components we consider in this ablation study contributes positively to overall performance.

The component that has the largest impact in performance is removing kernel regularization. We were not able to \emph{completely} remove regularization (i.e., use $\varepsilon = 0$ in \cref{eqn:regularized_poisson}), as doing so resulted in training failures in all scenes because of numerical errors (undefined values). Instead, we resorted to using a small cutoff in the denominator of the Poisson kernel in \cref{eqn:poisson}---an approach termed ``desingularization'' by \citet{cortez2001method}. In addition to significantly worsening quantitative scores in \cref{tab:ablation}, using desingularization results in extracted meshes with strong artifacts, similar to those we show in \cref{fig:regularized} for the unoptimized mesh.

\begin{table}[t]
	\centering
	\caption{Chamfer distances on BlendedMVS for ablation study. Labels indicate components we \emph{remove} from the full method we evaluate in \cref{tab:blendedmvs}.}
	\vspace{-0.75em}
	\begin{tabularx}{\linewidth}{@{} X >{\centering\arraybackslash}X >{\centering\arraybackslash}X >{\centering\arraybackslash}X >{\centering\arraybackslash}X >{\centering\arraybackslash}X >{\centering\arraybackslash}X >{\centering\arraybackslash}X @{}}
		\toprule
		\xmark & entropy loss & winding loss & normal loss & kernel reg. & point grow. & normal train. \\
		\midrule
		avg. & 0.66 & 0.65 & 0.69 & 0.73 & 0.64 & 0.68 \\
		\bottomrule
	\end{tabularx}
	\label{tab:ablation}
	\vspace{-1em}
\end{table}

\subsection{Rendering with shadow rays}\label{sec:shadows}

Compared to other fast point-based methods such as Gaussian surfels \citep{Dai2024GaussianSurfels}, our method uses ray tracing instead of rasterization. Ray tracing provides greater flexibility than rasterization in terms of rendering algorithms and light transport effects it can be used for. A salient example in the context of 3D reconstruction is rendering direct illumination via shadow rays \citep{ling2023shadowneus} when reconstructing scenes with known illumination---doing so is not possible with rasterization methods. We use a synthetic example to demonstrate that this additional flexibility translates to improvements in both mesh reconstruction and novel view synthesis.

\paragraph{Experiment setup} We re-render the \textsc{Lego} scene from the NeRF Realistic Synthetic dataset \citep{mildenhall2021nerf} with Lambertian materials and illumination from two point light sources. We render 200 images from random viewpoints and point-light positions that vary from image to image. We process these images with COLMAP to extract an initial dense point cloud, normals, and camera poses---we use ground truth point light positions for each view.

\begin{figure}
	\centering
	\includegraphics[width=\linewidth]{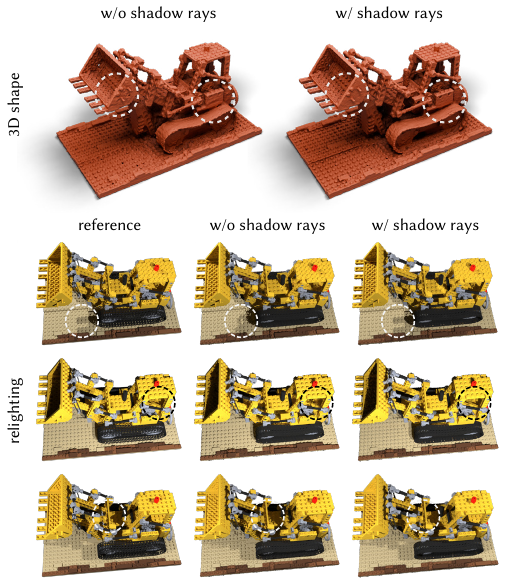}
	\caption{Comparison of extracted meshes and rendered images from training with and without shadow rays on the NeRF Realistic Synthetic \textsc{Lego} scene. Optimizing with shadow rays results in extracted meshes that have fewer artifacts and finer details. Additionally, it results in images rendered under novel lighting that have more accurate shadows.}
	\label{fig:shadow}
	\vspace{-1em}
\end{figure}

\paragraph{Inverse rendering} We optimize this initialization using our method with and without shadow rays. Without shadow rays, our method works exactly as before, using the radiance field representation of \cref{eqn:radiance_field} to model global (direct and indirect) illumination.

With shadow rays, we augment \cref{eqn:radiance_field} to include direct illumination terms for the two point light sources:
\begin{align}\label{eqn:radiance_field_shadow_rays}
	\radiance_{\text{sh.rays}}\paren{\point, \direction} &\equiv \sum_{i=1,2}\nolimits \radiance_{\mathrm{d}}^{i}\paren{\point, \direction} \nonumber \\
	&+ \mlp\paren{\point, \direction, \impnormal\paren{\point}, \regpcdl{\radfeat}^1\paren{\point}, \dots, \regpcdl{\radfeat}^K\paren{\point}},
\end{align}
where for each light source:
\begin{equation}\label{eqn:direct}
	\radiance_{\mathrm{d}}^{i}\paren{\point, \direction} \equiv \albedo\paren{\point} \transmittance\paren{\point,\light_i} \frac{\impnormal\paren{\point} \cdot \paren{\light_i-\point}}{\norm{\light_i-\point}^3}.
\end{equation}
Here, $\light_i$ is the position of the $i$-th light source, $\albedo$ is the albedo at $\point$, and $\transmittance\paren{\point,\light_i}$ is the exponential transmittance between $\point$ and $\light_i$. We compute albedo as an additional output of the MLP, and transmittance using quadrature (\cref{eqn:quadrature}) and fast queries.


\paragraph{Results} We compare in \cref{fig:shadow} extracted meshes and images rendered under novel lighting, after optimizing with and without shadow rays. We observe that optimizing with shadow rays results in extracted meshes with fewer artifacts and finer details. Additionally, the corresponding images have accurate shadows, compared to clearly implausible shadows otherwise. These results demonstrate that our method benefits from the generality of ray tracing, while achieving efficiency comparable to rasterization.

\section{Limitations and discussion}\label{sec:discussion}

We introduced the regularized dipole sum, a point-based representation for inverse rendering of 3D geometry. This representations allows modeling, ray tracing, and optimizing both implicit geometry and radiance fields using point cloud attributes. Coupled with Barnes-Hut acceleration, dipole sums enable multi-view 3D reconstruction at speeds comparable to and reconstruction quality better than rasterization methods, while maintaining the generality afforded by ray tracing. Starting from dense structure-from-motion initialization, dipole sums additionally produce surface reconstructions of better quality than neural representations, while escaping overfitting issues or computational overheads those encounter. We conclude with a discussion of some limitations of our work, and the future research directions they suggest.


\paragraph{Dealing with specular appearance} Both our work and prior work studying representations other than neural (e.g., Gaussian surfels \citep{Dai2024GaussianSurfels} for point-based, and Voxurf \citep{wu2022voxurf} for grid-based) report difficulties producing accurate surface reconstructions in areas of strong specular appearance. Our method and Voxurf alleviate the issue using shallow MLPs to predict appearance from interpolated features, which inevitably introduces a computational overhead. Gaussian surfels instead rely on data-driven priors, which in turn introduces reliance on supervised training and generalizability issues. Previous work on neural representations showed improved handling of specular appearance through the use of ``roughness'' \citep{verbin2022ref} or ``anisotropy'' \citep{miller2023theory} features that are combined with spherical-harmonic radiance representations during ray tracing. As our method also uses ray tracing, it could adapt this approach by incorporating such features as point attributes rather than neural network outputs.

\begin{figure}
	\centering
	\includegraphics[width=\linewidth]{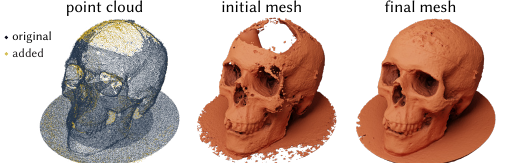}
	\caption{Visualization of original and added points on the point cloud for the DTU \textsc{skull} scene (left), and extracted meshes from the original (middle) and final (right) point clouds. Our point-growing method fills in regions with large gaps in the point cloud (e.g., top of the skull) and, together with optimized geometry attributes, fixes these gaps in the final extracted mesh.}
	\label{fig:growing}
	\vspace{-1em}
\end{figure}

\paragraph{Dealing with large textureless regions} Our method directly uses the dense point cloud from structure-from-motion initialization, which makes it sensitive to artifacts such as large holes and missing surfaces in that point cloud (e.g., due to textureless regions where structure from motion fails). Our method mitigates these artifacts through the use of learnable per-point geometry attributes (\cref{sec:moment}) and point growing (\cref{sec:implementation}), but the resulting reconstruction of very large textureless regions can still be noisy---\cref{fig:growing} shows an example. Adopting more elaborate point growing procedures from prior work \citep{xu2022point,kerbl20233d} could enable our technique to more effective mitigate such artifacts.

\paragraph{Global illumination and surface rendering} Our dipole sum representation is designed for efficient ray tracing. Thus, it is compatible, in principle, with more general (primal and differentiable) rendering algorithms. We have demonstrated this compatibility only in a restricted fashion, through a combination of dipole sums with shadow rays for direct illumination estimation (\cref{sec:experiments}). Additionally, we focused on volume rendering, but our dipole sum representation is also compatible with surface rendering formulations, which lead to improved surface reconstruction \citep{cai2022physics,luan2021unified} at the cost of needing to account for visibility discontinuities in the represented implicit surface \citep{vicini2022differentiable,bangaru2022differentiable}. In the future, it would be interesting to investigate combinations of dipole sums with other direct illumination \citep{bitterli2020spatiotemporal} and global illumination \citep{pharr2023physically} algorithms, in both volume and surface rendering formulations.

\paragraph{Applications beyond 3D reconstruction} We evaluated our point-based representation only in the narrow context of inverse rendering for 3D reconstruction. However, representations such as our dipole sum---comprising a tailored \emph{combination} of point cloud attributes, an interpolation kernel, and fast summation queries---can be useful more broadly for a variety of graphics and vision tasks, analogously to multiresolution hashgrids \citep{muller2022instant}. Broader adoption could be facilitated by investigation of alternative fast summation methods \citep{beatson1997short}, and data-driven optimization of interpolation kernels \citep{chen2023neurbf,ryan2022fast}.


\begin{acks}
	We thank Keenan Crane, Rohan Sawhney, and Nicole Feng for many helpful discussions, and the authors of \citet{Dai2024GaussianSurfels,neus2,li2023neuralangelo} for help running experimental comparisons. This work was supported by NSF award 1900849, NSF Graduate Research Fellowship DGE2140739, an NVIDIA Graduate Fellowship for Miller, and a Sloan Research Fellowship for Gkioulekas.
\end{acks}

\bibliographystyle{ACM-Reference-Format}
\bibliography{dipoles}

\appendix

\section{Proofs}

We prove the two propositions we presented in \cref{sec:regularized}.

\subsection{Proof of Proposition~\ref{pro:psr}}\label{app:psr}

Poisson surface reconstruction computes a scalar field as the solution to the following Poisson equation \citep[Section 3]{kazhdan2006poisson}:%
\footnote{The minus sign at the right-hand side is because we use outward normals, whereas \citet{kazhdan2006poisson} use inward ones.}
\begin{equation}\label{eqn:psr_bvp}
	\Delta \solution\paren{\point} = -\nabla \cdot \nfunc\paren{\point},\quad \point \in \R^3,
\end{equation}
where the \emph{normal field} $\nfunc : \R^3 \to \R^3$ in the right-hand side equals:
\begin{equation}\label{eqn:psr_source}
	\nfunc\paren{\point} \equiv \sum_{m=1}^M \phi_{\varepsilon}\paren{\point - \pc_m} \pcarea_m \pcnormal_m,
\end{equation}
and $\phi_{\varepsilon}$ is the Gaussian function in \cref{eqn:nascent}. As the domain of \cref{eqn:psr_bvp} is $\R^3$, which is unbounded:
\begin{enumerate*}
	\item existence of a solution requires that the right-hand side term decays sufficiently fast, which is true for $\nabla \cdot \nfunc$ thanks to the Gaussians in \cref{eqn:psr_source};
	\item uniqueness of that solution requires imposing a condition at infinity, and as $\solution$ should approximate the indicator in \cref{eqn:gauss}, the appropriate condition is that $\abs{\solution} \to 0$ as $\norm{\point} \to \infty$.
\end{enumerate*}
Under these conditions, the solution of \cref{eqn:psr_bvp} equals the \emph{Newtonian potential} with moment $\nabla \cdot \nfunc$ \citep[Section 2.2.1.b, Theorem 1]{evans2022partial}:%
%
%
\begin{equation}
	\solution\paren{\point} = -\int_{\R^3} \green\paren{\point, \pointalt} \nabla_{\pointalt} \cdot \nfunc\paren{\pointalt} \ud \pointalt.
\end{equation}
From \cref{eqn:psr_source}, this solution becomes:
\begin{equation}\label{eqn:psr_solution}
	\solution\paren{\point} = -\sum_{m=1}^M \pcarea_m \overbrace{\int_{\R^3} \green\paren{\point, \pointalt} \nabla_{\pointalt} \cdot \phi_{\varepsilon}\paren{\pointalt - \pc_m} \pcnormal_m \ud \pointalt}^{\equiv \subpot_m\paren{\point}}.
\end{equation}
We consider each of the $M$ integrals separately. Denoting by $\ball\paren{\point,\radius}\subset \R^3$ the ball with center $\point$ and radius $\radius$, we have:
\begin{align}
	\subpot_m\paren{\point} &= \lim_{\radius\to\infty}\int_{\ball\paren{\point,\radius}} \green\paren{\point, \pointalt} \nabla_{\pointalt} \cdot \phi_{\varepsilon}\paren{\pointalt - \pc_m} \pcnormal_m \ud \pointalt \label{eqn:psrproof1} \\
	&= \lim_{\radius\to\infty}\Bigg\{\int_{\partial\ball\paren{\point,\radius}} \green\paren{\point, \pointalt} \phi_{\varepsilon}\paren{\pointalt - \pc_m} \pcnormal_m \cdot \nicefrac{\pointalt - \point}{\radius} \surfMeasure\paren{\pointalt} \nonumber \\
	&\quad\quad\quad- \int_{\ball\paren{\point,\radius}} \nabla_{\pointalt} \green\paren{\point, \pointalt} \cdot \pcnormal_m \phi_{\varepsilon}\paren{\pointalt - \pc_m} \ud \pointalt\Bigg\} \label{eqn:psrproof2} \\
	&= 0 + \int_{\R^3} \nabla_{\point} \green\paren{\point, \pointalt} \cdot \pcnormal_m \phi_{\varepsilon}\paren{\pointalt - \pc_m} \ud \pointalt \label{eqn:psrproof3} \\
	&= \nabla_{\point} \green_{\varepsilon}\paren{\point-\pc_m} \cdot \pcnormal_m \label{eqn:psrproof4} \\
	&= - \poisson_{\varepsilon}\paren{\point,\pc_m}. \label{eqn:psrproof5}
\end{align}
In this sequence: \labelcref{eqn:psrproof1} reexpresses the unbounded integration domain; \labelcref{eqn:psrproof2} uses integration by parts; \labelcref{eqn:psrproof3} uses the distributive property of limits and the facts that $\green\paren{\point,\pointalt} \phi_{\varepsilon}\paren{\pointalt-\pc_m} = o(\norm{\pointalt-\point}^{-2})$ and $\nabla_{\pointalt} \green\paren{\point, \pointalt} = - \nabla_{\point} \green\paren{\point, \pointalt}$; \labelcref{eqn:psrproof4} follows from the definition in \labelcref{eqn:green_regularized} and the properties of the Green's function; and \labelcref{eqn:psrproof5} follows from the definition in \labelcref{eqn:poisson_regularized}. Then, from \cref{eqn:psr_solution,eqn:psrproof5,eqn:regularized_pcwn},
\begin{equation}
	\solution\paren{\point} = \sum_{m=1}^M \pcarea_m \poisson_{\varepsilon}\paren{\point,\pc_m} = \regpcwn\paren{\point}.
\end{equation}
This concludes our proof. We note two differences with the numerical implementation of PSR by \citet{kazhdan2006poisson}:
\begin{enumerate}[leftmargin=*]
	\item To make the Poisson equation \labelcref{eqn:psr_bvp} amenable to a linear-system solver, \citet{kazhdan2006poisson} impose Dirichlet boundary conditions on a bounding volume of the point cloud. For the true indicator function in \cref{eqn:gauss}, these conditions and our condition that $\solution \to 0$ at infinity are equivalent. However, for point-cloud approximations, they are not equivalent and the choice between them is arbitrary \citep[Section 4.4]{kazhdan2013screened}.
	\item \Citet{kazhdan2006poisson} suggest variable per-point standard deviations $\varepsilon_m$. \Cref{pro:psr} still holds in that case, except using $\varepsilon_m$ in \cref{eqn:regularized_pcwn}. We comment on this suggestion in \cref{sec:moment}.
\end{enumerate}

\subsection{Proof of Proposition~\ref{pro:expected_winding_number}}\label{app:noise}

Under the assumptions of \cref{pro:expected_winding_number}, we have from \cref{eqn:pcwn}:
\begin{align}
	\!\!\!\!\Exp{\curly{\pcrandom_m,\pcrandomnormal_m}_{m=1}^M\!\!}{\pcwn\paren{\point}}\! 
	&= \sum_{m=1}^M \pcarea_m \Exp{\pcrandom_m,\pcrandomnormal_m}{\poisson\paren{\point, \pcrandom_m}} \label{eqn:proof2} \\
	&= \sum_{m=1}^M \pcarea_m \cdot\Exp{\pcrandom_m}{\Exp{\pcrandomnormal_m}{\conditional{\poisson\paren{\point, \pcrandom_m}}{\pcrandom_m}}} \label{eqn:proofm1} \\
	&= \sum_{m=1}^M \pcarea_m \Exp{\pcrandom_m}{\Exp{\pcrandomnormal_m}{\conditional{\pcrandomnormal_m\nabla\!\green\paren{\point, \pcrandom_m}}{\pcrandom_m}}} \label{eqn:proofm2} \\
	&= \sum_{m=1}^M \pcarea_m \Exp{\pcrandom_m}{\impfeat_m\pcnormal_m\nabla\!\green\paren{\point, \pcrandom_m}} \label{eqn:proofm3} \\
	&= \sum_{m=1}^M \pcarea_m \pcnormal_m\nabla\,\Exp{\pcrandom_m}{\green\paren{\point, \pcrandom_m}}\impfeat_m \label{eqn:proof4} \\
	&= \sum_{m=1}^M \pcarea_m \pcnormal_m\nabla\!\green_\varepsilon\paren{\point, \pc_m} \impfeat_m\label{eqn:proof5} \\
	&= \sum_{m=1}^M \pcarea_m \poisson_\varepsilon\paren{\point, \pc_m}\impfeat_m  \label{eqn:proof6} \\
	&= \regpcdl{\impfeat}. \label{eqn:proof7}
\end{align}
In this sequence: \labelcref{eqn:proof2} follows from linearity of expectation; \labelcref{eqn:proofm1} follows from the law of total expectation; \labelcref{eqn:proofm2} follows from the definition in \labelcref{eqn:poisson}; \labelcref{eqn:proofm3} follows from the assumptions on $\pcrandomnormal_m$; \labelcref{eqn:proof4} follows from the fact that differentiation and expectation commute; \labelcref{eqn:proof6} follows from the definition in \labelcref{eqn:poisson_regularized}; and \labelcref{eqn:proof7} follows from the definition in \cref{eqn:dipole_sum}. The only non-trivial step is \labelcref{eqn:proof5}. From the assumption that $\pcrandom_m$ is a Gaussian random variable, we have (up to a constant scale that we omit for simplicity):
\begin{align}
	\Exp{\pcrandom_m}{\green\paren{\point, \pcrandom_m}} &\propto \int_{\pointalt\in\R^3} \green\paren{\point, \pointalt} \exp\paren{-\frac{\norm{\point - \pointalt}^2}{2 \varepsilon^2}} \ud \pointalt \\
	&\propto \int_{\pointalt\in\R^3} \green\paren{\point, \pointalt} \phi_\varepsilon\paren{\point - \pointalt} \ud \pointalt \label{eqn:proof9} \\
	&= \green_\varepsilon\paren{\point, \pointalt}, \label{eqn:proof10}
\end{align}
where \labelcref{eqn:proof9} follows from the definition in \labelcref{eqn:nascent}. The step \labelcref{eqn:proof10} follows from the fact that \labelcref{eqn:proof7} is equivalent, by the properties of the Green's function, to the solution of the partial differential equation in \labelcref{eqn:green_regularized}.

\section{Entropy loss}\label{app:entropy}

The \emph{free-flight distribution} \citep{miller2023theory} of a ray $\ray_{\camera,\view}\paren{\distancealt}$,
\begin{equation}
	p^{\mathrm{ff}}_{\camera,\view}\paren{\distancealt} \equiv \exp\paren{-\int_{0}^\distancealt \coeff\paren{\ray_{\camera,\view}\paren{\distance},\view} \ud \distance}\coeff\paren{\ray_{\camera,\view}\paren{\distancealt},\view},
\end{equation}
is the probability density function for a first intersection occurring at $\distancealt$. For surface-like volumes, the free-flight distribution should approximate a Dirac delta. We can encourage such behavior by penalizing the Shannon entropy of the free-flight distribution along each ray---low entropy favors peaked unimodal distributions. To do so, we use quadrature (\cref{eqn:quadrature}) to form a discrete approximation of the free-flight distribution at the ray samples $\distancealt_\mathrm{n} = \distancealt_0 < \dots < \distancealt_J = \distancealt_\mathrm{f}$:

\begin{equation}\label{eqn:quadrature_pff}
	p_j  \equiv \exp\paren{-\sum_{i=1}^{j} \coeff_{i}\Delta_{i}} \paren{1 - \exp\paren{\coeff_{j}\Delta_{j}}}.
\end{equation}
We then compute the Shannon entropy of the vector $\bracket{p_1,\dots,p_J}$,
\begin{equation}\label{eqn:shannon_entropy}
	H\paren{\camera,\view} \equiv - \sum_{j = 1}^J p_j \log p_j.
\end{equation}
We accumulate such entropies for all rays in the loss $\loss_{\mathrm{entropy}}$.

\section{Backpropagation details}\label{app:backprop}

As in \cref{sec:backprop}, throughout this section we use $\dirichlet$ as a stand-in for any of the \emph{moment attributes} stored in $\pcloud$---namely, the geometry attribute $\impfeat$ and the appearance attributes $\radfeat^k,\ k=1,\dots,K$. As we discuss in \cref{sec:backprop_details}, in practice we implement the backpropagation operations in \cref{eqn:farfield_gradient,eqn:farfield_dipole_features} for all these attributes as vector operations updating all attributes in parallel.

\paragraph{Backpropagation to nodes} An adjoint query backpropagates a derivative $\dd\regpcdld\paren{\point}$---provided by differentiable rendering---to all tree nodes that contributed to this dipole sum during the corresponding primal query. At each such node $t$, the query increments the (vector-valued) gradient attribute $\dd\nodedirichlet_t$ by an amount that follows from differentiating \cref{eqn:farfield}:
\begin{equation}\label{eqn:farfield_gradient}
	\nodearea_t \regul\paren{\frac{\norm{\nodepc_t-\point}}{\varepsilon}}\frac{\nodepc_t-\point}{\norm{\nodepc_t-\point}^3} \cdot \dd\regpcdld\paren{\point}.
\end{equation}

\paragraph{Second-stage backpropagation to leaf nodes} This stage backpropagates accumulated gradient attributes $\dd\nodedirichlet_t$ from all nodes to leaf nodes corresponding to individual points $\pc_m, m=1,\dots,M$ in $\pcloud$. For each such leaf node, we denote by $\AS\paren{m}$ the set of its ancestor nodes in the tree. Then, by differentiating \cref{eqn:farfield_dipole_features}, we can express this backpropagation stage as simply:
\begin{equation}\label{eqn:farfield_features_gradient}
	\dd\!\dirichlet_m = \sum_{t \in \AS\paren{m}} \frac{\pcarea_m}{\nodearea_t} \pcnormal_m \dd\nodedirichlet_t.
\end{equation}
Each leaf node has $\bigO\paren{\log M}$ ancestors, thus total complexity of the second stage is $\bigO\paren{M \log M}$. In practice we implement \cref{eqn:farfield_features_gradient} as a matrix-vector multiplication that has negligible cost.

\end{document}